\documentstyle[theapa,jair,twoside,url,11pt]{article}

\jairheading{12}{2000}{317--337}{6/99}{5/00}
\ShortHeadings{Axiomatizing Causal Structures}{Halpern}
\firstpageno{317}

\begin{document}
\input{psfig}

\newtheorem{THEOREM}{Theorem}[section]
\newenvironment{theorem}{\begin{THEOREM} \hspace{-.85em} {\bf :} }%
                        {\end{THEOREM}}
\newtheorem{LEMMA}[THEOREM]{Lemma}
\newenvironment{lemma}{\begin{LEMMA} \hspace{-.85em} {\bf :} }%
                      {\end{LEMMA}}
\newtheorem{COROLLARY}[THEOREM]{Corollary}
\newenvironment{corollary}{\begin{COROLLARY} \hspace{-.85em} {\bf :} }%
                          {\end{COROLLARY}}
\newtheorem{PROPOSITION}[THEOREM]{Proposition}
\newenvironment{proposition}{\begin{PROPOSITION} \hspace{-.85em} {\bf :} }%
                            {\end{PROPOSITION}}
\newtheorem{DEFINITION}[THEOREM]{Definition}
\newenvironment{definition}{\begin{DEFINITION} \hspace{-.85em} {\bf :} \rm}%
                            {\end{DEFINITION}}
\newtheorem{CLAIM}[THEOREM]{Claim}
\newenvironment{claim}{\begin{CLAIM} \hspace{-.85em} {\bf :} \rm}%
                            {\end{CLAIM}}
\newtheorem{EXAMPLE}[THEOREM]{Example}
\newenvironment{example}{\begin{EXAMPLE} \hspace{-.85em} {\bf :} \rm}%
                            {\end{EXAMPLE}}
\newtheorem{REMARK}[THEOREM]{Remark}
\newenvironment{remark}{\begin{REMARK} \hspace{-.85em} {\bf :} \rm}%
                            {\end{REMARK}}

\newcommand{\thm}{\begin{theorem}}
\newcommand{\lem}{\begin{lemma}}
\newcommand{\pro}{\begin{proposition}}
\newcommand{\dfn}{\begin{definition}}
\newcommand{\rem}{\begin{remark}}
\newcommand{\xam}{\begin{example}}
\newcommand{\cor}{\begin{corollary}}
\newcommand{\prf}{\noindent{\bf Proof:} }
\newcommand{\ethm}{\end{theorem}}
\newcommand{\elem}{\end{lemma}}
\newcommand{\epro}{\end{proposition}}
\newcommand{\edfn}{\bbox\end{definition}}
\newcommand{\erem}{\bbox\end{remark}}
\newcommand{\exam}{\bbox\end{example}}
\newcommand{\ecor}{\end{corollary}}
\newcommand{\eprf}{\bbox\vspace{0.1in}}
\newcommand{\beqn}{\begin{equation}}
\newcommand{\eeqn}{\end{equation}}
\newcommand{\wbox}{\mbox{$\sqcap$\llap{$\sqcup$}}}
\newcommand{\bbox}{\vrule height7pt width4pt depth1pt}
\newcommand{\qed}{\eprf}
\newcommand{\clm}{\begin{claim}}
\newcommand{\eclm}{\end{claim}}
\let\member=\in
\let\notmember=\notin
\newcommand{\sub}{_}
\def\su{^}
\newcommand{\rarrow}{\rightarrow}
\newcommand{\larrow}{\leftarrow}
\newcommand{\boldsymbol}[1]{\mbox{\boldmath $\bf #1$}}
\newcommand{\bolda}{{\bf a}}
\newcommand{\boldb}{{\bf b}}
\newcommand{\boldc}{{\bf c}}
\newcommand{\boldd}{{\bf d}}
\newcommand{\bolde}{{\bf e}}
\newcommand{\boldf}{{\bf f}}
\newcommand{\boldg}{{\bf g}}
\newcommand{\boldh}{{\bf h}}
\newcommand{\boldi}{{\bf i}}
\newcommand{\boldj}{{\bf j}}
\newcommand{\boldk}{{\bf k}}
\newcommand{\boldl}{{\bf l}}
\newcommand{\boldm}{{\bf m}}
\newcommand{\boldn}{{\bf n}}
\newcommand{\boldo}{{\bf o}}
\newcommand{\boldp}{{\bf p}}
\newcommand{\boldq}{{\bf q}}
\newcommand{\boldr}{{\bf r}}
\newcommand{\bolds}{{\bf s}}
\newcommand{\boldt}{{\bf t}}
\newcommand{\boldu}{{\bf u}}
\newcommand{\boldv}{{\bf v}}
\newcommand{\boldw}{{\bf w}}
\newcommand{\boldx}{{\bf x}}
\newcommand{\boldy}{{\bf y}}
\newcommand{\boldz}{{\bf z}}
\newcommand{\boldA}{{\bf A}}
\newcommand{\boldB}{{\bf B}}
\newcommand{\boldC}{{\bf C}}
\newcommand{\boldD}{{\bf D}}
\newcommand{\boldE}{{\bf E}}
\newcommand{\boldF}{{\bf F}}
\newcommand{\boldG}{{\bf G}}
\newcommand{\boldH}{{\bf H}}
\newcommand{\boldI}{{\bf I}}
\newcommand{\boldJ}{{\bf J}}
\newcommand{\boldK}{{\bf K}}
\newcommand{\boldL}{{\bf L}}
\newcommand{\boldM}{{\bf M}}
\newcommand{\boldN}{{\bf N}}
\newcommand{\boldO}{{\bf O}}
\newcommand{\boldP}{{\bf P}}
\newcommand{\boldQ}{{\bf Q}}
\newcommand{\boldR}{{\bf R}}
\newcommand{\boldS}{{\bf S}}
\newcommand{\boldT}{{\bf T}}
\newcommand{\boldU}{{\bf U}}
\newcommand{\boldV}{{\bf V}}
\newcommand{\boldW}{{\bf W}}
\newcommand{\boldX}{{\bf X}}
\newcommand{\boldY}{{\bf Y}}
\newcommand{\boldZ}{{\bf Z}}
\newcommand{\sat}{\models}
\newcommand{\dtur}{\models}
\newcommand{\infers}{\vdash}
\newcommand{\stur}{\vdash}
\newcommand{\rimp}{\Rightarrow}
\newcommand{\limp}{\Leftarrow}
\newcommand{\dimp}{\Leftrightarrow}
\newcommand{\bor}{\bigvee}
\newcommand{\band}{\bigwedge}
\newcommand{\union}{\cup}
\newcommand{\inter}{\cap}
\newcommand{\xx}{{\bf x}}
\newcommand{\yy}{{\bf y}}
\newcommand{\uu}{{\bf u}}
\newcommand{\vv}{{\bf v}}
\newcommand{\FF}{{\bf F}}
\newcommand{\natnum}{{\sl N}}
\newcommand{\IR}{\mbox{$I\!\!R$}}
\newcommand{\IP}{\mbox{$I\!\!P$}}
\newcommand{\IN}{\mbox{$I\!\!N$}}
\newcommand{\IC}{\mbox{$C\!\!\!\!\raisebox{.75pt}{\mbox{\sqi I}}$}}
\newcommand{\marrow}{\hbox{$\rightarrow$ \hskip -10pt
                      $\rightarrow$ \hskip 3pt}}
\renewcommand{\phi}{\varphi}
\newcommand{\Circ}{\mbox{{\small $\bigcirc$}}}
\newcommand{\lt}{<}
\newcommand{\gt}{>}
\newcommand{\all}{\forall}
\newcommand{\infinity}{\infty}
\newcommand{\bc}[2]{\left( \begin{array}{c} #1 \\ #2  \end{array} \right)}
\newcommand{\cross}{\times}
\newcommand{\bigfootnote}[1]{{\footnote{\normalsize #1}}}
\newcommand{\medfootnote}[1]{{\footnote{\small #1}}}
\newcommand{\bd}{\bf}


\newcommand{\imp}{\Rightarrow}

\newcommand{\A}{{\cal A}}
\newcommand{\B}{{\cal B}}
\newcommand{\C}{{\cal C}}
\newcommand{\D}{{\cal D}}
\newcommand{\E}{{\cal E}}
\newcommand{\F}{{\cal F}}
\newcommand{\G}{{\cal G}}
\newcommand{\I}{{\cal I}}
\newcommand{\J}{{\cal J}}
\newcommand{\K}{{\cal K}}
\newcommand{\M}{{\cal M}}
\newcommand{\N}{{\cal N}}
\newcommand{\Ocal}{{\cal O}}
\newcommand{\Hcal}{{\cal H}}
\renewcommand{\P}{{\cal P}}
\newcommand{\Q}{{\cal Q}}
\newcommand{\R}{{\cal R}}
\newcommand{\T}{{\cal T}}
\newcommand{\U}{{\cal U}}
\newcommand{\V}{{\cal V}}
\newcommand{\W}{{\cal W}}
\newcommand{\X}{{\cal X}}
\newcommand{\Y}{{\cal Y}}
\newcommand{\Z}{{\cal Z}}

\newcommand{\Kone}{{\cal K}_1}
\newcommand{\abs}[1]{\left| #1\right|}
\newcommand{\set}[1]{\left\{ #1 \right\}}
\newcommand{\Ki}{{\cal K}_i}
\newcommand{\Kn}{{\cal K}_n}
\newcommand{\st}{\, \vert \,} 
\newcommand{\stc}{\, : \,} 
\newcommand{\la}{\langle}
\newcommand{\ra}{\rangle}
\newcommand{\<}{\langle}
\renewcommand{\>}{\rangle}
\newcommand{\lang}{\mbox{${\cal L}_n$}}
\newcommand{\langd}{\mbox{${\cal L}_n^D$}}

\newtheorem{nlem}{Lemma}
\newtheorem{Ob}{Observation}
\newtheorem{pps}{Proposition}
\newtheorem{defn}{Definition}
\newtheorem{crl}{Corollary}
\newtheorem{cl}{Claim}
\newcommand{\pf}{\par\noindent{\bf Proof}~~}
\newcommand{\eg}{e.g.,~}
\newcommand{\ie}{i.e.,~}
\newcommand{\vs}{vs.~}
\newcommand{\cf}{cf.~}
\newcommand{\etal}{et al.\ }
\newcommand{\resp}{resp.\ }
\newcommand{\respc}{resp.,\ }
\newcommand{\comment}[1]{\marginpar{\scriptsize\raggedright #1}}
\newcommand{\wrt}{with respect to~}
\newcommand{\re}{r.e.}
\newcommand{\nind}{\noindent}
\newcommand{\distributed}{distributed\ }
\newcommand{\bn}{\bigskip\markright{NOTES}
\section*{Notes}}
\newcommand{\Exer}{
\bigskip\markright{EXERCISES}
\section*{Exercises}}
\newcommand{\DG}{D_G}
\newcommand{\Sm}{{\rm S5}_m}
\newcommand{\Smc}{{\rm S5C}_m}
\newcommand{\Smi}{{\rm S5I}_m}
\newcommand{\Smic}{{\rm S5CI}_m}
\newcommand{\Martin}{Mart\'\i n\ }
\newcommand{\ol}{\setlength{\itemsep}{0pt}\begin{enumerate}}
\newcommand{\eol}{\end{enumerate}\setlength{\itemsep}{-\parsep}}
\newcommand{\ul}{\setlength{\itemsep}{0pt}\begin{itemize}}
\newcommand{\dl}{\setlength{\itemsep}{0pt}\begin{description}}
\newcommand{\edl}{\end{description}\setlength{\itemsep}{-\parsep}}
\newcommand{\eul}{\end{itemize}\setlength{\itemsep}{-\parsep}}
\newtheorem{fthm}{Theorem}
\newtheorem{flem}[fthm]{Lemma}
\newtheorem{fcor}[fthm]{Corollary}
\newcommand{\slidehead}[1]{
\eject
\Huge
\begin{center}
{\bf #1 }
\end{center}
\vspace{.5in}
\LARGE}

\newcommand{\subG}{_G}
\newcommand{\If}{{\bf if}}

\newcommand{\attime}{{\tt \ at\_time\ }}
\newcommand{\hatell}{\skew6\hat\ell\,}
\newcommand{\Then}{{\bf then}}
\newcommand{\Until}{{\bf until}}
\newcommand{\Else}{{\bf else}}
\newcommand{\Repeat}{{\bf repeat}}
\newcommand{\cA}{{\cal A}}
\newcommand{\cE}{{\cal E}}
\newcommand{\cF}{{\cal F}}
\newcommand{\cI}{{\cal I}}
\newcommand{\cN}{{\cal N}}
\newcommand{\cR}{{\cal R}}
\newcommand{\cS}{{\cal S}}
\newcommand{\BN}{B^{\scriptscriptstyle \cN}}
\newcommand{\BS}{B^{\scriptscriptstyle \cS}}
\newcommand{\cW}{{\cal W}}
\newcommand{\EG}{E_G}
\newcommand{\CG}{C_G}
\newcommand{\CN}{C_\cN}
\newcommand{\ES}{E_\cS}
\newcommand{\EN}{E_\cN}
\newcommand{\CS}{C_\cS}

\newcommand{\attack}{\mbox{{\it attack}}}
\newcommand{\attacking}{\mbox{{\it attacking}}}
\newcommand{\delivered}{\mbox{{\it delivered}}}
\newcommand{\exist}{\mbox{{\it exist}}}
\newcommand{\decide}{\mbox{{\it decide}}}
\newcommand{\clean}{{\it clean}}
\newcommand{\diff}{{\it diff}}
\newcommand{\Failed}{{\it failed}}
\newcommand\eqdef{=_{\rm def}}
\newcommand{\true}{\mbox{{\it true}}}
\newcommand{\false}{\mbox{{\it false}}}

\newcommand{\DN}{D_{\cN}}
\newcommand{\DS}{D_{\cS}}
\newcommand{\tyme}{{\it time}}
\newcommand{\fp}{f}

\newcommand{\Kax}{{\rm K}_n}
\newcommand{\Kaxc}{{\rm K}_n^C}
\newcommand{\Kaxd}{{\rm K}_n^D}
\newcommand{\Tax}{{\rm T}_n}
\newcommand{\Taxc}{{\rm T}_n^C}
\newcommand{\Taxd}{{\rm T}_n^D}
\newcommand{\fourax}{{\rm S4}_n}
\newcommand{\fouraxc}{{\rm S4}_n^C}
\newcommand{\fouraxd}{{\rm S4}_n^D}
\newcommand{\fiveax}{{\rm S5}_n}
\newcommand{\fiveaxc}{{\rm S5}_n^C}
\newcommand{\fiveaxd}{{\rm S5}_n^D}
\newcommand{\Dax}{{\rm KD45}_n}
\newcommand{\Daxc}{{\rm KD45}_n^C}
\newcommand{\Daxd}{{\rm KD45}_n^D}
\newcommand{\LP}{{\cal L}_n}
\newcommand{\LCP}{{\cal L}_n^C}
\newcommand{\LDP}{{\cal L}_n^D}
\newcommand{\LCDP}{{\cal L}_n^{CD}}
\newcommand{\MP}{{\cal M}_n}
\newcommand{\MPr}{{\cal M}_n^r}
\newcommand{\MPrt}{\M_n^{\mbox{\scriptsize{{\it rt}}}}}
\newcommand{\MPrst}{\M_n^{\mbox{\scriptsize{{\it rst}}}}}
\newcommand{\MPelt}{\M_n^{\mbox{\scriptsize{{\it elt}}}}}
\renewcommand{\lang}{\mbox{${\cal L}_{n} (\Phi)$}}
\renewcommand{\langd}{\mbox{${\cal L}_{n}^D (\Phi)$}}
\newcommand{\fiveaxdu}{{\rm S5}_n^{DU}}
\newcommand{\LPD}{{\cal L}_n^D}
\newcommand{\fiveaxu}{{\rm S5}_n^U}
\newcommand{\fiveaxcu}{{\rm S5}_n^{CU}}
\newcommand{\LPU}{{\cal L}^{U}_n}
\newcommand{\LPCU}{{\cal L}_n^{CU}}
\newcommand{\LDPU}{{\cal L}_n^{DU}}
\newcommand{\LCPU}{{\cal L}_n^{CU}}
\newcommand{\LPDU}{{\cal L}_n^{DU}}
\newcommand{\LPCDU}{{\cal L}_n^{\it CDU}}
\newcommand{\Cn}{\C_n}
\newcommand{\CSnp}{\I_n^{oa}(\Phi')}
\newcommand{\CSc}{\C_n^{oa}(\Phi)}
\newcommand{\Ccs}{\C_n^{oa}}
\newcommand{\CSAX}{OA$_{n,\Phi}$}
\newcommand{\CSAXN}{OA$_{n,{\Phi}}'$}
\newcommand{\untill}{U}
\newcommand{\until}{\, U \,}
\newcommand{\amp}{{\rm a.m.p.}}
\newcommand{\commentout}[1]{}
\newcommand{\msgc}[1]{ @ #1 }
\newcommand{\Camp}{{\C_n^{\it amp}}}
\newcommand{\bi}{\begin{itemize}}
\newcommand{\ei}{\end{itemize}}
\newcommand{\be}{\begin{enumerate}}
\newcommand{\ee}{\end{enumerate}}
\newcommand{\rarrowr}{\stackrel{r}{\rightarrow}}
\newcommand{\ack}{\mbox{\it ack}}
\newcommand{\Gz}{\G_0}
\newcommand{\denselist}{\itemsep 0pt\partopsep 0pt}
\def\seealso#1#2{({\em see also\/} #1), #2}
\newcommand{\cents}{\hbox{\rm \rlap{/}c}}

\newenvironment{oldthm}[1]{\par\noindent{\bf Theorem #1:} \em \noindent}{\par}
\newenvironment{oldlem}[1]{\par\noindent{\bf Lemma #1:} \em \noindent}{\par}
\newenvironment{oldcor}[1]{\par\noindent{\bf Corollary #1:} \em \noindent}{\par}
\newenvironment{oldpro}[1]{\par\noindent{\bf Proposition #1:} \em \noindent}{\par}
\newcommand{\othm}[1]{\begin{oldthm}{\ref{#1}}}
\newcommand{\eothm}{\end{oldthm} \medskip}
\newcommand{\olem}[1]{\begin{oldlem}{\ref{#1}}}
\newcommand{\eolem}{\end{oldlem} \medskip}
\newcommand{\ocor}[1]{\begin{oldcor}{\ref{#1}}}
\newcommand{\eocor}{\end{oldcor} \medskip}
\newcommand{\opro}[1]{\begin{oldpro}{\ref{#1}}}
\newcommand{\eopro}{\end{oldpro} \medskip}

\newcommand{\world}{W}
\newcommand{\WN}{\W_N}
\newcommand{\Winf}{\W^*}
\newcommand{\tends}{\rightarrow}
\newcommand{\tendsto}{\tends}
\newcommand{\ninfty}{{N \rightarrow \infty}}
\newcommand{\nworldsv}[1]{{\it \#worlds}^{#1}}
\newcommand{\nworlds}{{{\it \#worlds}}_{N}^{\epsvec}}
\newcommand{\nwrldPnt}[1]{\nworlds[#1]}
\newcommand{\nworldsarg}[1]{\nworlds[#1]}
\newcommand{\binco}[2]{{{#1}\choose{#2}}}
\newcommand{\closure}[1]{{\overline{#1}}}
\newcommand{\balpha}{\bar{\alpha}}
\newcommand{\bbeta}{\bar{\beta}}
\newcommand{\bgamma}{\bar{\gamma}}
\newcommand{\half}{\frac{1}{2}}
\newcommand{\bQ}{\overline{Q}}
\newcommand{\Vector}[1]{{\langle #1 \rangle}}
\newcommand{\Algzeroone}{\mbox{\em Compute01}}
\newcommand{\Algcompute}{\mbox{{\em Compute-Pr}$_\infty$}}

\newcommand{\Prinfv}[1]{{\Pr}^{#1}_\infty}
\newcommand{\PrNv}[1]{{\Pr}^{#1}_N}
\newcommand{\Prinf}{{\Pr}_\infty}
\newcommand{\PrN}{{\Pr}_N}
\newcommand{\pN}{\PrN (\phi | \KB)}
\newcommand{\IPrinf}{{\Box\Diamond\Prinf}}
\newcommand{\PPrinf}{{\Diamond\Box\Prinf}}
\newcommand{\prNw}{{\Pr}_{N}^{\epsvec}}
\newcommand{\prNwi}{{\Pr}_{N^i}^{\epsvec^i}}
\newcommand{\priw}{{\Pr}_{\infty}}
\newcommand{\beliefprob}[1]{\Pr(#1)} 
\newcommand{\Prinfeps}{{{\Pr}_\infty^{\epsvec}}}
\newcommand{\PrNeps}{{{\Pr}_N^{\epsvec}}}
\newcommand{\Pinf}[2]{{\Pi_\infty^{#1}[#2]}}

\newcommand{\infvocab}{\Omega}
\newcommand{\infunvocab}{\Upsilon}
\newcommand{\vocab}{\Phi}
\newcommand{\nonunaryvocab}{\vocab}
\newcommand{\unvocab}{\Psi}

\newcommand{\cL}{{\cal L}}
\newcommand{\cLne}{{\cal L}^-}
\newcommand{\finitelang}{\cL_i^d(\Phi)}
\newcommand{\fLd}{\cL_d^d(\Phi)}
\newcommand{\cLd}{\cL_{\mbox{\em \scriptsize$d$}}(\Phi)}
\newcommand{\Laeq}{{\cal L}^{\aeq}}
\newcommand{\Leq}{{\cal L}^{=}}
\newcommand{\Lunaeq}{\Laeq_1}
\newcommand{\Luneq}{\Leq_1}

\newcommand{\KB}{{\it KB}}
\newcommand{\phieven}{\phi_{\mbox{\footnotesize\it even}}}
\newcommand{\phiodd}{\phi_{\mbox{\footnotesize\it odd}}}
\newcommand{\KBM}{\KB_{\boldM}}
\newcommand{\Pstar}{P^*}
\newcommand{\barP}{\neg P}
\newcommand{\barQ}{\neg Q}
\newcommand{\maxarity}{\rho}
\newcommand{\Init}{\mbox{\it Init\/}}
\newcommand{\Rep}{\mbox{\it Rep\/}}
\newcommand{\Acc}{\mbox{\it Acc\/}}
\newcommand{\Comp}{\mbox{\it Comp\/}}
\newcommand{\Step}{\mbox{\it Step\/}}
\newcommand{\Univ}{\mbox{\it Univ\/}}
\newcommand{\Exis}{\mbox{\it Exis\/}}
\newcommand{\Between}{\mbox{\it Between\/}}
\newcommand{\Count}{\mbox{\it count\/}}
\newcommand{\phiq}{\phi_Q}
\newcommand{\propform}{\beta}
\newcommand{\allphi}{\ast}
\newcommand{\ID}{\mbox{\it ID}}
\newcommand{\atomdesc}{{\psi_*}}
\newcommand{\rigid}{\mbox{\it rigid}}
\newcommand{\abdesc}{\widehat{D}}
\newcommand{\KBtwo}{\theta}
\newcommand{\psits}{\psi[\KBtwo,\abdesc]}
\newcommand{\psitsnohat}{\psi[\KBtwo,D]}
\newcommand{\KBfly}{\KB_{\mbox{\scriptsize \it fly}}}
\newcommand{\KBchirps}{\KB_{\mbox{\scriptsize \it chirps}}}
\newcommand{\KBmagpie}{\KB_{\mbox{\scriptsize \it magpie}}}
\newcommand{\KBhep}{\KB_{\mbox{\scriptsize \it hep}}}
\newcommand{\KBnixon}{\KB_{\mbox{\scriptsize \it Nixon}}}
\newcommand{\KBel}{\KB_{\mbox{\scriptsize \it likes}}}
\newcommand{\KBtax}{\KB_{\mbox{\scriptsize \it taxonomy}}}
\newcommand{\KBarm}{\KB_{\mbox{\scriptsize \it arm}}}
\newcommand{\KBlate}{\KB_{\mbox{\scriptsize \it late}}}
\newcommand{\KBp}{{\KB'}}
\newcommand{\KBflyp}{\KB_{\mbox{\scriptsize \it fly}}'}
\newcommand{\KBpp}{{\KB''}}
\newcommand{\KBdef}{\KB_{\mbox{\scriptsize \it def}}}
\newcommand{\KBdishep}{\KB_{\mbox{\scriptsize \it $\lor$hep}}}

\newcommand{\canKB}{\widehat{\KB}}
\newcommand{\canxi}{\widehat{\xi}}
\newcommand{\KBfo}{\KB_{\it fo}}
\newcommand{\KBconst}{\psi}
\newcommand{\KBprop}{\KBp}

\newcommand{\quak}{\mbox{\it Quaker\/}}
\newcommand{\repub}{\mbox{\it Republican\/}}
\newcommand{\pac}{\mbox{\it Pacifist\/}}
\newcommand{\Nixon}{\mbox{\it Nixon\/}}
\newcommand{\Winged}{\mbox{\it Winged\/}}
\newcommand{\Winner}{\mbox{\it Winner\/}}
\newcommand{\Child}{\mbox{\it Child\/}}
\newcommand{\Boy}{\mbox{\it Boy\/}}
\newcommand{\Tall}{\mbox{\it Tall\/}}
\newcommand{\Elephant}{\mbox{\it Elephant\/}}
\newcommand{\Gray}{\mbox{\it Gray\/}}
\newcommand{\Yellow}{\mbox{\it Yellow\/}}
\newcommand{\Clyde}{\mbox{\it Clyde\/}}
\newcommand{\Tweety}{\mbox{\it Tweety\/}}
\newcommand{\Opus}{\mbox{\it Opus\/}}
\newcommand{\Bird}{\mbox{\it Bird\/}}
\newcommand{\Penguin}{{\it Penguin\/}}
\newcommand{\Fish}{\mbox{\it Fish\/}}
\newcommand{\Fly}{\mbox{\it Fly\/}}
\newcommand{\Warmblooded}{\mbox{\it Warm-blooded\/}}
\newcommand{\White}{\mbox{\it White\/}}
\newcommand{\Red}{\mbox{\it Red\/}}
\newcommand{\Giraffe}{\mbox{\it Giraffe\/}}
\newcommand{\Visible}{\mbox{\it Easy-to-see\/}}
\newcommand{\Bat}{\mbox{\it Bat\/}}
\newcommand{\Blue}{\mbox{\it Blue\/}}
\newcommand{\Fever}{\mbox{\it Fever\/}}
\newcommand{\Jaun}{\mbox{\it Jaun\/}}
\newcommand{\Hep}{\mbox{\it Hep\/}}
\newcommand{\Eric}{\mbox{\it Eric\/}}
\newcommand{\Alice}{\mbox{\it Alice\/}}
\newcommand{\Tom}{\mbox{\it Tom\/}}
\newcommand{\Lottery}{\mbox{\it Lottery\/}}
\newcommand{\Zookeeper}{\mbox{\it Zookeeper\/}}
\newcommand{\Fred}{\mbox{\it Fred\/}}
\newcommand{\Likes}{\mbox{\it Likes\/}}
\newcommand{\Day}{\mbox{\it Day\/}}
\newcommand{\Nextday}{\mbox{\it Next-day\/}}
\newcommand{\Sleepslate}{\mbox{\it To-bed-late\/}}
\newcommand{\Riseslate}{\mbox{\it Rises-late\/}}
\newcommand{\TS}{\mbox{\it TS\/}}
\newcommand{\EEJ}{\mbox{\it EEJ\/}}
\newcommand{\FC}{\mbox{\it FC\/}}
\newcommand{\Dodo}{\mbox{\it Dodo\/}}
\newcommand{\Ab}{\mbox{\it Ab\/}}
\newcommand{\Chirps}{\mbox{\it Chirps\/}}
\newcommand{\Swims}{\mbox{\it Swims\/}}
\newcommand{\Magpie}{\mbox{\it Magpie\/}}
\newcommand{\Moody}{\mbox{\it Moody\/}}
\newcommand{\SomeMorning}{\mbox{\it Tomorrow\/}}
\newcommand{\Animal}{\mbox{\it Animal\/}}
\newcommand{\Sparrow}{\mbox{\it Sparrow\/}}
\newcommand{\Turtle}{\mbox{\it Turtle\/}}
\newcommand{\Older}{\mbox{\it Over60\/}}
\newcommand{\Patient}{\mbox{\it Patient\/}}
\newcommand{\Black}{\mbox{\it Black\/}}
\newcommand{\Ray}{\mbox{\it Ray\/}}
\newcommand{\Reiter}{\mbox{\it Reiter\/}}
\newcommand{\Drew}{\mbox{\it Drew\/}}
\newcommand{\McDermott}{\mbox{\it McDermott\/}}
\newcommand{\Emu}{\mbox{\it Emu\/}}
\newcommand{\Canary}{\mbox{\it Canary\/}}
\newcommand{\BlueCanary}{\mbox{\it BlueCanary\/}}
\newcommand{\FlyingBird}{\mbox{\it FlyingBird\/}}
\newcommand{\UL}{\mbox{\it LeftUsable\/}}
\newcommand{\UR}{\mbox{\it RightUsable\/}}
\newcommand{\BL}{\mbox{\it LeftBroken\/}}
\newcommand{\BR}{\mbox{\it RightBroken\/}}
\newcommand{\Ticket}{\mbox{\it Ticket\/}}
\newcommand{\BlueEyed}{{\mbox{\it BlueEyed\/}}}
\newcommand{\Jaundice}{{\it Jaundice\/}}
\newcommand{\Hepatitis}{{\it Hepatitis\/}}
\newcommand{\HeartDisease}{{\mbox{\it Heart-disease\/}}}
\newcommand{\bJ}{{\overline{J}\,}}
\newcommand{\bH}{{\overline{H}\,}}
\newcommand{\bB}{{\overline{B}\,}}
\newcommand{\Prem}{{\mbox{\it Child\/}}}
\newcommand{\David}{{\mbox{\it David\/}}}
\newcommand{\Son}{{\mbox{\it Son\/}}}	

\newcommand{\ceslim}{Ces\`{a}ro limit}
\newcommand{\Sigmad}{\Sigma^d_i}
\newcommand{\Liogonkii}{Liogon'ki\u\i}
\newcommand{\Vstar}{{\modfrag_*}}
\newcommand{\moddesc}{\psi \land \modfrag}
\newcommand{\sumact}{\Degr_2}
\newcommand{\degree}{\Degr_1}
\newcommand{\Active}{\alpha}
\newcommand{\ActiveAtoms}{\boldA}
\newcommand{\named}{n}
\newcommand{\aactive}{a}
\newcommand{\degr}{\delta}
\newcommand{\chsize}{f}
\newcommand{\chconst}{g}
\newcommand{\Chconst}{G}
\newcommand{\Degr}{\Delta}
\newcommand{\frags}{\M}
\newcommand{\const}{H}
\newcommand{\modfrag}{\V}
\newcommand{\AD}{\A}
\newcommand{\Named}{\nu}
\newcommand{\bit}{b}
\newcommand{\guess}{\gamma}
\newcommand{\bxor}[1]{\dot{\bor}}
\newcommand{\weight}{\omega}
\newcommand{\arity}{{\it arity}}
\newcommand{\assigned}{\leftarrow}
\newcommand{\snum}[2]{{{#1} \brace {#2}}}

\newcommand{\eps}{\tau}
\newcommand{\vareps}{\varepsilon}
\newcommand{\varepsvec}{{\vec{\vareps}}}
\newcommand{\xtuple}{\vec{x}}
\newcommand{\ctuple}{{\vec{c}\,}}
\newcommand{\uvec}{{\!{\vec{\,u}}}}
\newcommand{\ucom}{u}
\newcommand{\pvec}{{\!{\vec{\,p}}}}
\newcommand{\pcom}{p}
\newcommand{\vvec}{\vec{v}}
\newcommand{\wvec}{\vec{w}}
\newcommand{\xvec}{\vec{x}}
\newcommand{\yvec}{\vec{y}}
\newcommand{\zvec}{\vec{z}}
\newcommand{\zerovec}{\vec{0}}

\newcommand{\epscom}{\eps}
\newcommand{\epsvec}{{\vec{\eps}\/}}
\newcommand{\prop}[2]{{||{#1}||_{{#2}}}}
\newcommand{\aeq}{\approx} 
\newcommand{\app}{\approx}
\newcommand{\alt}{\prec}
\newcommand{\aleq}{\preceq}
\newcommand{\agt}{\succ}
\newcommand{\ageq}{\succeq}
\newcommand{\altne}{\prec}
\newcommand{\naeq}{\not\approx}
\newcommand{\cprop}[3]{{\|{#1}|{#2}\|_{{#3}}}}
\newcommand{\Bigcprop}[3]{{\Bigl\|{#1}\Bigm|{#2}\Bigr\|_{{#3}}}}

\newcommand{\reals}{\IR}
\newcommand{\qsep}{\,}
\newcommand{\perm}{\pi}
\newcommand{\val}{V}
\newcommand{\rwent}{\mbox{$\;|\!\!\!\sim$}_{\mbox{\scriptsize \it rw}}\;}
\newcommand{\notrwent}{\mbox{$\;|\!\!\!\not\sim$}_{\mbox{\scriptsize\it rw}}\;}
\newcommand{\dempster}{\delta}
\newcommand{\dentails}{{\;|\!\!\!\sim\;}}
\newcommand{\notdentails}{{\;|\!\!\!\not\sim\;}}
\newcommand{\dentailssub}[1]{\dentails\hspace{-0.4em}_{
          \mbox{\scriptsize{\it #1}}}\;}
\newcommand{\notdentailssub}[1]{\notdentails\hspace{-0.4em}_{
          \mbox{\scriptsize{\it #1}}}\;}
\newcommand{\default}{\rightarrow}

\newcommand{\vecof}[1]{{\pi({#1})}}
\newcommand{\PIN}[2]{{\Pi_N^{#1}[#2]}}
\newcommand{\SS}[2]{{S^{#1}[#2]}}
\newcommand{\SSc}[2]{{S^{#1}[#2]}}
\newcommand{\SSzero}[1]{\SS{\zerovec}{#1}}
\newcommand{\SSczero}[1]{\SSc{\zerovec}{#1}}
\newcommand{\SSpos}[1]{\SS{\leq \zerovec}{#1}}
\newcommand{\Sol}{{\it Sol}}
\newcommand{\poscon}{\gamma}
\newcommand{\constraints}{\Gamma}
\newcommand{\constpos}{\constraints^{\leq}}
\newcommand{\propspace}{\Delta^K}
\newcommand{\mept}{\vvec}
\newcommand{\mecoord}{v}
\newcommand{\mepts}{{\Q}}
\newcommand{\OS}{{\cal O}}
\renewcommand{\S}{{\cal S}}
\newcommand{\meval}{{\rho}}
\newcommand{\meptmin}{{\uvec^{\ast}_{\mbox{\scriptsize min}}}}
\newcommand{\sizeof}[1]{{\sigma(#1)}}
\newcommand{\mesize}{{\sigma^{\ast}}}
\renewcommand{\pf}{\alpha}
\newcommand{\ps}{\beta}
\newcommand{\Atoms}{\AD}
\newcommand{\limNstar}{{\lim_{N \rightarrow \infty}}^{\!\!\!*}\:}
\newcommand{\probf}[2]{F_{[#1|#2]}}
\newcommand{\probfun}[1]{F_{[#1]}}
\newcommand{\Prmu}[1]{{\mu}_{#1}}
\newcommand{\foversion}[1]{\xi_{#1}}
\newcommand{\fly}{\mbox{\it fly\/}}
\newcommand{\bird}{\mbox{\it bird\/}}
\newcommand{\yellow}{\mbox{\it yellow\/}}
\newcommand{\propconsts}{\Lambda}
\newcommand{\alldiff}{\chi^{\neq}}
\newcommand{\unaryD}{D^1}
\newcommand{\nonunD}{D^{> 1}}
\newcommand{\eqD}{D^{=}}

\newcommand{\AX}{\mbox{AX}}
\newcommand{\AXrec}{\mbox{AX}_{\rm rec}}
\newcommand{\AXun}{\mbox{AX}_{\rm uniq}}
\newcommand{\AXex}{\mbox{AX}^+}
\newcommand{\AXrecex}{\mbox{AX}^+_{\rm rec}}
\newcommand{\AXunex}{\mbox{AX}^+_{\rm uniq}}
\newcommand{\Trec}{{\cal T}_{\rm rec}}
\newcommand{\Tun}{{\cal T}_{\rm uniq}}
\newcommand{\Lprop}{{\cal L}_{\rm uniq}}
\newcommand{\LGP}{{\cal L}_{\rm GP}}
\newcommand{\Lex}{{\cal L}^+}
\newcommand{\Pre}{\mbox{Pre}}

\title{Axiomatizing Causal Reasoning}
\author{\name Joseph Y.\ Halpern \email halpern@cs.cornell.edu\\
\addr   Cornell University, Computer Science Department\\
   Ithaca, NY 14853\\
   http://www.cs.cornell.edu/home/halpern
}
\maketitle

\begin{abstract}
Causal models defined in terms of a collection of equations, as
defined by Pearl, are axiomatized here.  Axiomatizations are provided
for three successively more general classes of causal models: (1) the
class of recursive theories (those without feedback), (2) the class of
theories where the solutions to the equations are unique, (3)
arbitrary theories (where the equations may not have solutions
and, if they do, they are not necessarily unique).  It is shown
that to reason about causality in the most general third class,
we must extend the language used by Galles and Pearl
\citeyear{GallesPearl97,GallesPearl98}.  
In addition, the complexity of the decision procedures is characterized
for all the languages and classes of models considered.
\end{abstract}

\section{Introduction}
The important role of causal reasoning---in prediction, explanation, and
counterfactual reasoning---has been argued eloquently in a number of
recent papers and books
\cite{CH97,HeckShac,DH90,DruzdzelSimon93,Pearl.Biometrika,PearlVerma91,SpirtesSG}.
If we are to reason
about causality, then it is certainly useful to find axioms that
characterize such reasoning.  The way we go about axiomatizing causal
reasoning depends on two critical factors:
\begin{itemize}
\item how we model causality, and
\item the language that we use to reason about it.
\end{itemize}

In this paper, I consider one approach to modeling causality,
using {\em structural equations}.
The use of structural
equations as a model for causality is standard in the social sciences,
and seems to go back to the work of Sewall Wright in the 1920s (see
\cite{Goldberger72} for a discussion);
the particular framework that I use here
is due to Pearl \citeyear{Pearl.Biometrika}.  Galles and Pearl
\citeyear{GallesPearl97} introduce some axioms
for causal reasoning in this framework; 
they also provide a complete axiomatic characterization of reasoning about
causality
in this framework, under the strong assumption that there is a fixed,
given {\em causal ordering\/} $\prec$ of the equations \cite{GallesPearl98}.
Roughly speaking, this means there is a way
of ordering the variables that appear in the equations and we have
explicit axioms that say $X_j$ has no influence of $X_i$ if $X_i
\prec X_j$ in this causal ordering.

In this paper, I extend the results of Galles and Pearl by providing a
complete axiomatic
characterization for three increasingly general classes of causal
models (defined by structural equations):
\begin{enumerate}
\item the class of recursive theories (those without feedback---this
generalizes the situation considered by Galles and Pearl
\citeyear{GallesPearl98}, since every fixed causal ordering of the
variables gives rise to a recursive theory),
\item the class of
theories where the solutions to the equations are unique,
\item arbitrary theories (where the equations may not have solutions
and, if they do, they are not necessarily unique).
\end{enumerate}
In the process, I clarify some problems in the Galles-Pearl completeness
proof that arise from the lack of propositional connectives
(particularly disjunction) in
the language they consider and, more generally, highlight the role of
the language in reasoning about causality.
I also characterize the complexity of the decision problem for all these
languages and classes of models.

The rest of the paper is organized as follows.  In Section~\ref{syntax},
I give the syntax and semantics of the languages I will be considering
and review the definition of modifiable causal models.  In
Section~\ref{axiomatizations}, I present the complete axiomatizations.
In Section~\ref{decisionp} I consider the complexity of the decision
procedure.
I conclude in Section~\ref{conclusion}.

\section{Syntax and Semantics}\label{syntax}
An axiomatization is given with respect to a  particular language and a
class of models, so we need to make both precise.  Both the language and
models I use are based on those considered by Galles and Pearl
\citeyear{GallesPearl97,GallesPearl98}.  To make comparisons easier, I
use their notation as much as possible.  I start with the semantic
model, since it motivates some of the choices in the syntax, then give
the syntax, and finally define the semantics of formulas.

\subsection{Causal Models}
The basic picture here is that we are interested in the values of random
variables, some of which have a causal effect on others.  This effect
is modeled by a set of {\em structural equations}.

In practice, it
seems useful to split the random variables into two sets, the {\em
exogenous\/} variables, whose values are determined by factors outside
the model, and the {\em endogenous\/} variables.  It is these endogenous
variables whose values are described by the structural equations.

More
formally, a {\em signature\/} $\S$ is a tuple $(\U,\V,\R\}$,
where $\U$ is a finite set of exogenous variables, $\V$ is a finite set
of endogenous variables, and $\R$ associates with every variable $Y \in
\U \union \V$ a nonempty set $\R(Y)$
of possible values for $Y$ (the {\em range\/} of possible values of
$Y$).  Unless explicitly noted otherwise, I assume that $\R(Y)$ is a
{\em finite\/} set for each $Y \in \U \union \V$ and $|\R(Y)| \ge 2$.
The assumption that $\U$ and $\V$ are finite is relatively innocuous;
as we shall see, the assumption that $\R(Y)$ is finite has more of an
impact on the axioms. The assumption that $|\R(Y)| \ge 2$
allows us to ignore the trivial situation where $|\R(Y)| = 1$.  If $|\R(Y)|
=
1$, we can just remove the variable $Y$ from the signature without loss
of expressiveness.

A {\em causal model\/} over signature $\S$ is a tuple $T=(\S,\F)$
where $\F$ associates with each variable $X \in \V$ a function denoted
$F_X$ such that $F_X: (\times_{U \in \U} \R(U))
\times (\times_{Y \in \V - \{X\}} \R(Y)) \rightarrow \R(X)$.
$F_X$ tells us the value of $X$
given the values of all the other variables in $\U \union \V$.
We think of the functions $F_X$ as defining a set of {\em (modifiable)
structural equations}, relating the values of the variables.  Because
$F_X$ is a function, there is a unique value of $X$ once we have set all
the other variables.
Notice we have such functions only for the endogenous variables.
The exogenous variables
are taken as given; it is their effect on the endogenous
variables (and the effect of the endogenous variables on each other)
that we are modeling with the structural equations.

Given a causal model $T = (\S,\F)$ over signature $\S$, a (possibly
empty)  vector
$\vec{X}$ of variables in $\V$, and vectors $\vec{x}$ and
$\vec{u}$ of values for the variables in
$\vec{X}$ and $\U$, respectively, we can define a new causal model
denoted
$T_{\vec{X} \gets \vec{x}}(\vec{u})$ over the signature $\S_{\vec{X}}
= (\emptyset, \V - \vec{X}, \R|_{\V - \vec{X}})$.%
\footnote{I am implicitly identifying the vector $\vec{X}$ with the
subset of $\V$ consisting of the variables in $\vec{X}$.  I do this
throughout the paper.  $\R|_{\V - \vec{X}}$ is the restriction of $\R$
to the variables in $\V - \vec{X}$.}
Intuitively, this is the causal model that results when the variables in
$\vec{X}$ are set to $\vec{x}$
and the variables in $\U$ are set to $\vec{u}$.
Formally, $T_{\vec{X} \gets \vec{x}}(\vec{u}) = (\S_{\vec{X}},
\F^{\vec{X} \gets \vec{x},\vec{u}}\})$,
where $F_Y^{\vec{X} \gets \vec{x},\vec{u}}$ is obtained from $F_Y$
by setting the values of the
variables in $\vec{X}$ to $\vec{x}$ and the values of the variables
in $\U$ to $\vec{u}$.  The causal model
$T_{\vec{X} \gets \vec{x}}(\vec{u})$ is called a {\em submodel\/} of $T$
by Pearl \citeyear{pearl:99}.  It can describe a possible {\em
counterfactual\/} situation; that is, even though, under normal
circumstances, setting the exogenous variables to $\vec{u}$ may result
in the variables $\vec{X}$ having values $\vec{x}' \ne \vec{x}$, this
submodel describes what happens if they are set to $\vec{x}$ due to
some ``external action'', the cause of which is not modeled explicitly.
For example, to determine if the manufacturer is at fault in an accident
that involved a poorly maintained car, we may want to consider what
would have happened had the car been well maintained.  If there is a
random variable in the signature that describes how well maintained the
car is, then this means examining the submodel where that random
variable is set to 1 (the car is well maintained).  It is this ability
to examine counterfactual situations that makes causal structures a
useful tool for reasoning about causality.

Notice that, in general, there may not be a unique vector of values that
simultaneously satisfies the equations in $T_{\vec{X} \gets
\vec{x}}(\vec{u})$; indeed, there may not be a solution at all.  One
special case where there is guaranteed to be such a unique solution is
if there is some total ordering $\prec$ of the variables in $\V$ such
that if $X \prec Y$, then $F_X$ is independent of the value of $Y$; \ie
$F_X(\ldots, y, \ldots) = F_X(\ldots, y', \ldots)$ for all $y, y' \in
\R(Y)$.  In this case, the causal model is said to be {\em
recursive\/} or {\em acyclic\/}.  Intuitively, if the
theory is recursive, there is no
feedback.  If $X \prec Y$, then the value of $X$ may affect the value of
$Y$, but the value of $Y$ has no effect on the value of $X$.

It should be clear that if $T$ is a recursive theory, then there is
always a unique solution to the equations in
$T_{\vec{X} \gets \vec{x}}(\vec{u})$, for all $\vec{X}$, $\vec{x}$, and
$\vec{u}$.  (We simply solve for the variables in the order given by
$\prec$.)  On the other hand, as the following example shows, it is not
hard to construct nonrecursive theories for
which there is always a unique solution to the equations that arise.

\xam Let $\S = (\emptyset, \{X,Y\}, \R\})$, where
$\R(X) = \R(Y) =
\{-1, 0, 1\}$, and let $T = (\S,\F)$, where
$F_X$ is characterized by the equation $X = Y$ and $F_Y$
is characterized by the equation $Y = -X$ (that is, $F_X(y) = y$ and
$F_Y(x) = -x$). Clearly $T$ is not
recursive; the value of $X$ depends on the value of $Y$ and the value
of $Y$ depends on that of $X$. Nevertheless, it is easy to see that
$T$ has the unique solution $X=0,Y=0$, $T_{X \gets x}$ has the unique
solution $Y=-x$, and $T_{Y \gets y}$ has the unique solution $X = y$.
\exam

In this paper, I consider three successively more general classes of
causal models for a given signature $\S = (\U,\V,\R)$.
\begin{enumerate}
\item $\Trec(\S)$: the class of recursive causal models over
signature $\S$,
\item $\Tun(\S)$: the class of causal models $T$ over $\S$
where for all $\vec{X} \subseteq \V$, $\vec{x}$,
and $\vec{u}$, the equations in $T_{\vec{X} \gets \vec{x}}(\vec{u})$
have a unique solution,
\item $\T(\S)$: the class of all causal models over $\S$.
\end{enumerate}
I often omit the signature $\S$ when it is clear from
context or irrelevant, but the reader should bear in mind its
important role.

Why should we be interested in causal models that do not possess unique
solutions?   Are there real causal systems that do not possess unique
solutions?  The issue of whether nonrecursive system
can be given a causal interpretation is discussed at some
length by Strotz and Wold \citeyear{SW60}.  They argue that there are
reasonable ways of interpreting causal interpretations where the answer
is yes.  Under these interpretations, there
may well be more than one solution to the equations.  Perhaps the best
way to view such equations is to think of the
variables in $\V$ as being mutually interdependent; changing any
one of them may cause a change in the others.  (Think of demand and
supply in
economics or populations of rabbits and wolves.)  The solutions to the
equations then represent equilibrium situations.  If there is more than
one equilibrium, there will be more than one solution to the equations.
Of course, if there are no equilibria, then there will be no solutions
to the equations.

A related way of thinking about these equations is that they represent
atemporal versions of temporal causal equations.  That is, suppose that
we replace every variable $Y \in \U \union \V$ by a family of variables
$Y_0, Y_1, Y_2, \ldots$, where, intuitively, $Y_t$ represents the value
of $Y$ at time $t$.  Each equation $f_{X} \in \F$ is then replaced by a
family of equations $f_{X_t}$, where $f_{X_t}$ depends only on exogenous
variables $U_{t'}$ with $t' \le t$ and endogenous variables $Y_{t'}$
with $t' < t$.  This gives us a recursive system.  The values of $X_t$
under some setting of the variables with subscript 0 represents the
evolution of $X$ under that setting of the variables.  If $X_t$
eventually stabilizes, then we might expect the equilibrium value to be
the value of $X$ in some solution to the original set of equations.  If
$X_t$ stabilizes, then there would not be a solution to the original set
of equations.

\subsection{Syntax}  I focus here on two languages.  Both languages are
parameterized by a signature $\S$.
The first language, $\Lex(\S)$, borrows ideas from dynamic logic
\cite{Har}.  Again, I often write $\Lex$ rather than $\Lex(\S)$
(and similarly for the other languages defined below) to simplify the
notation. A {\em basic causal formula\/} is one of the form
$[Y_1 \gets y_1, \ldots, Y_k \gets y_k] \phi$, where $\phi$ is a Boolean
combination of formulas of the form $X(\vec{u}) = x$, $Y_1,
\ldots, Y_k, X$ are variables in $\V$, $Y_1, \ldots, Y_k$ are
distinct, $x \in \R(X)$, and $\vec{u}$ is a vector of values
for all the variables in $\U$.   I typically abbreviate such a formula
as $[\vec{Y} \gets \vec{y}]\phi$.
The special case where $k=0$ (which
is allowed) is abbreviated as $[\true]\phi$.
$[\vec{Y} \gets \vec{y}] X(\vec{u}) = x$ can be interpreted as ``in all
possible solutions to the structural equations obtained after setting
$Y_i$ to $y_i$, $i = 1, \ldots, k$, and the exogenous variables to
$\vec{u}$, random variable $X$ has value $x$''. As we shall see,
this formula is true in a causal model
$T$ if in all solutions to the equations in
$T_{\vec{Y} \gets \vec{y}}(\vec{u})$, the random variable $X$ has value
$x$.  Note that this formula is trivially true if there are no
solutions to the structural equations.  A {\em causal formula\/} is a
Boolean combination of basic causal formulas.

Just as with dynamic logic, we
can also define the formula $\<\vec{Y} \gets
\vec{y}\>(X(\vec{u}) = x)$ to be an abbreviation of
$\neg [\vec{Y} \gets \vec{y}] \neg (X(\vec{u}) = x)$.
$\<\vec{Y} \gets \vec{y}\> (X(\vec{u}) = x)$ is the dual of
$[\vec{Y} \gets \vec{y}] (X(\vec{u}) = x)$; it is true if, in some
solution to the structural equations obtained after setting $Y_i$ to
$y_i$, $i = 1, \ldots, k$, and the exogenous variables to $\vec{u}$,
random variable $X$ has value $x$.
 Taking $\true(\vec{u})$ to be an abbreviation for
$X(\vec{u})
= x
\lor
X(\vec{u}) \ne x$ for some variable $X$ and $x \in \R(X)$, and taking
$\false(\vec{u})$ to be an abbreviation for $\neg \true(\vec{u})$, we
have that $\<\vec{Y}
\gets \vec{y}\>\true(\vec{u})$ is true if there is some solution to the
equations obtained by setting $Y_i$ to $y_i$, $i = 1, \ldots, k$, and
the variables in $\U$ to $\vec{u}$
(since $[\vec{Y} \gets \vec{y}]\false(\vec{u})$ says that in every
solution to the equations obtained by setting $Y_i$ to $y_i$ and $\U$ to
$\vec{u}$, the formula
$\false(\vec{u})$ is true, and thus holds exactly if the equations
have no solution).

Let $\Lprop(\S)$ be the sublanguage of $\Lex(\S)$
which consists of Boolean combinations of
formulas of the form $[\vec{Y} \gets \vec{y}]X(\vec{u}) = x$.
Thus, the difference between $\Lprop$ and
$\Lex$ is that in $\Lprop$, only $X(\vec{u}) = x$ is allowed after
$[\vec{Y} \gets \vec{y}]$, while in $\Lex$, arbitrary Boolean
combinations of formulas of the form $X(\vec{u}) = x$ are allowed.
As we shall see, for reasoning about causality in $\Tun$, the language
$\Lprop$ is adequate, since it is equivalent in expressive power to
$\Lex$.  However, this is no longer the case when reasoning about
causality in $\T$.

Following Galles and Pearl's notation, I often write
$[\vec{Y} \gets \vec{y}]X(\vec{u}) = x$ as
$X_{\vec{Y} \gets \vec{y}}(\vec{u})
= x$. If $\vec{Y}$ is clear from context or
irrelevant, I further
abbreviate this as $X_{\vec{y}}(\vec{u}) = x$.  (This is actually the
notation used by Galles and Pearl.)
Let $\LGP(\S)$ be the sublanguage of $\Lprop(\S)$ consisting of just
conjunctions of formulas
of the form $X_{\vec{y}}(\vec{u}) = x$.  In particular, it does
not contain disjunctions or negations of such formulas.  Although Galles and
Pearl \citeyear{GallesPearl98} are not explicit about the
language they are using, it seems to be $\LGP$.%
\footnote{This was confirmed by Judea Pearl [private communication, 1997].}

\subsection{Semantics}
A formula in $\Lex(\S)$ is true or false in a
causal model in $\T(\S)$.  As usual,
we write $T \sat \phi$ if the causal formula $\phi$ is true in causal
model $T$.  For a basic causal formula, we have
$T \sat [\vec{Y} \gets \vec{y}](X(\vec{u}) = x)$ if in all solutions
to $T_{\vec{Y} \gets \vec{y}}(\vec{u})$ (\ie in all vectors of values
for the variables in $\V - \vec{Y}$ that  simultaneously satisfy all the
equations $F^{\vec{Y} \gets \vec{y}}_Z$, for $Z \in \V - \vec{Y}$), the
variable
$X$ has value
$x$.
We define the truth value of arbitrary causal formulas, which
are just Boolean combinations of basic causal formulas, in the obvious
way:
\begin{itemize}
\item $T \sat \phi_1 \land \phi_2$ if $T \sat \phi_1$ and $T \sat
\phi_2$
\item $T \sat  \neg\phi$ if $T \not\sat \phi$.
\end{itemize}

As usual, we say that a formula $\phi$ is {\em valid\/} with respect to
a class $\T'$ of causal models if $T \sat \phi$ for all $T \in \T'$.

I can now make precise the earlier claim that in $\Tun$ (and
hence $\Trec$), the language $\Lprop$ is just as expressive as the full
language $\Lex$.
\lem\label{uniquelem} The following formulas are valid in $\Tun$:
\begin{itemize}
\item[(a)] $\Tun \sat [\vec{Y} \gets \vec{y}](\phi \lor \psi) \dimp
[\vec{Y} \gets \vec{y}]\phi \lor [\vec{Y} \gets \vec{y}]\psi$,
\item[(b)] $\Tun \sat [\vec{Y} \gets \vec{y}](\phi \land \psi) \dimp
[\vec{Y} \gets \vec{y}]\phi \land [\vec{Y} \gets \vec{y}]\psi$,
\item[(c)] $\Tun \sat [\vec{Y} \gets \vec{y}]\neg \phi \dimp \neg
[\vec{Y} \gets \vec{y}] \phi$.
\end{itemize}
Hence, in $\Tun$, every formula in $\Lex$ is equivalent to a formula
in $\Lprop$.
\elem

\prf Straightforward; left to the reader.  \eprf

Note that it follows from these equivalences that in $\Tun$,
$[\vec{Y} \gets \vec{y}]\phi$ is equivalent to
$\<\vec{Y} \gets \vec{y}\>\phi$.  It is also worth noting that
Lemma~\ref{uniquelem}(b) holds in arbitrary
causal models in $\T$, not just in $\Tun$.
However, parts (a) and (c) do not, as the
following example shows.

\xam\label{nonrec}
Let $\S = (\emptyset, \{X,Y\}, \R)$, where
$\R(X) = \R(Y) = \{0, 1\}$; let $T = (\S,\F)$, where
$F_X$ is characterized by the equation $X = Y$ and
$F_Y$ is characterized by the equation $Y = X$.
Clearly $T \notin \Tun$; both $(0,0)$ and
$(1,1)$ are solutions to $T$.  It is easy to see
that $T \sat [\true](X = 0 \lor X=1) \land \neg[\true](X=0)
\land \neg [\true](X=1)$, showing that part (a)
of  Lemma~\ref{uniquelem} does not hold in $T$,
and that
$T \sat \neg[\true](X=1) \land
\neg[\true] \neg(X=1)$, showing that part (c)
does not hold either. \exam

\section{Complete Axiomatizations}\label{axiomatizations}

I briefly recall some standard definitions from logic.  An {\em axiom
system\/}\index{axiom
system}~AX consists of a collection of {\em axioms\/}\index{axiom|(}
and {\em inference rules}\index{inference rule|(}.
An axiom is a formula (in some predetermined language ${\cal L}$), and
an inference
rule has the form ``from $\phi_1, \ldots, \phi_k$ infer~$\psi$,''
where $\phi_1, \ldots, \phi_k, \psi$ are formulas in ${\cal L}$.
A {\em proof\/} in AX consists of a
sequence of formulas in ${\cal L}$, each of which is either an axiom
in~AX or follows by an application of an inference rule.
A proof is said to be a {\em proof of the formula~$\phi$} if the last
formula in the proof is~$\phi$.
We say~$\phi$ is {\em provable in~AX},
and write $\mbox{AX} \infers \phi$,
if there is a proof of~$\phi$ in~AX; similarly, we say that $\phi$ is
{\em consistent with AX\/} if $\neg \phi$ is not provable in AX.

An axiom system AX is said to be {\em sound\/} for
a language ${\cal L}$ with respect to a class
$\T'$ of causal models if every formula in ${\cal L}$
provable in AX is valid with respect to $\T'$.
AX is {\em complete\/} for
${\cal L}$ with respect to
${\cal T'}$ if every formula in ${\cal L}$
that is valid with respect to $\T'$ is
provable in AX.

We now want to find axioms that characterize the classes of causal
models in which we are interested, namely $\Trec$, $\Tun$, and $\T$.
To deal with $\Trec$, it is helpful to define $Y
\leadsto Z$, read ``$Y$ affects $Z$'', as an abbreviation for the
formula
$$\lor_{\vec{X}\subseteq \V, \vec{x} \in \times_{X \in \V}\R(X), y \in
\R(y),\vec{u} \in \times_{U \in \U} \R(U), z \ne z' \in \R(Z)}
(\mbox{$Z_{\vec{x}y}(\vec{u}) = z'$} \land
Z_{\vec{x}}(\vec{u}) = z).$$  Thus, $Y$ affects $Z$ if there is some
setting of the exogenous variables and
some other endogenous variables
for which changing the value of $Y$ changes the
value of $Z$.  This definition is used in axiom C6 below, which
characterizes recursiveness.

Consider the following axioms:
\begin{itemize}
\item[C0.] All instances of propositional tautologies.
\item[C1.] $X_{\vec{y}}(\vec{u}) = x \rimp X_{\vec{y}}(\vec{u}) \ne
x'$ if $x, x' \in \R(X), x \ne x'$
\hfill  (equality)
\item[C2.] $\lor_{x \in \R(X)} X_{\vec{y}}(\vec{u}) = x$
\hfill (definiteness)
\item[C3.] $(W_{\vec{x}}(\vec{u}) = w \land Y_{\vec{x}}(\vec{u}) =
y) \rimp Y_{\vec{x}w}(\vec{u}) = y$
\hfill (composition)
\item[C4.] $X_{x\vec{w}}(\vec{u}) = x$ \hfill (effectiveness)
\item[C5.] $(Y_{\vec{x}w}(\vec{u}) = y \land W_{\vec{x}y}(\vec{u}) = w)
\rimp Y_{\vec{x}}(\vec{u}) = y$
\hfill (reversibility)
\item[C6.] $(X_0 \leadsto X_1 \land \ldots \land X_{k-1} \leadsto X_k)
\rimp
\neg (X_k \leadsto X_0)$ \mbox{ } \hfill
(recursiveness)
\end{itemize}

We have one rule of inference:
\begin{itemize}
\item[MP.] From $\phi$ and $\phi \rimp \psi$, infer $\psi$ \hfill (modus
ponens)
\end{itemize}

C1 just states an obvious property of equality: if $X=x$ for
every solution of the
equations in $T_{\vec{x}}(\vec{u})$, then we cannot have $X = x'$, if
$x' \ne x$.%
\footnote{In an earlier draft of this paper, where C1 and C2 were
introduced, C1 was called ``uniqueness''.  Galles and Pearl
\citeyear{GallesPearl98} then adopted this name as well.  In retrospect,
this axiom really does not say anything about uniqueness.  The axiom
which does is D10, which will be discussed later.}
In a richer language, this could have been expressed as
$(X_{\vec{y}}(\vec{u}) = x \land X_{\vec{y}}(\vec{u}) = x') \rimp (x =
x')$, but this formula is not in $\Lex$ (since $\Lex$ does not include
expressions such as $x' = x$).
C2 states that there is some value $x
\in \R(X)$ which is the value of $X$ in all solutions to the equations in
$T_{\vec{x}}(\vec{u})$.  C2 is not valid in $\T$, but it is valid in
$\Tun$.  Note that in stating C2, I am making use of the fact that $\R(X)$
is
finite (otherwise C2 would involve an infinite disjunction, and would no
longer be a formula in $\Lprop$).  In fact,
it can be shown that
if we allow signatures where the sets $\R(X)$ are infinite, we
include C2 only for those random variables $X$ such that $\R(X)$ is
finite.%
\footnote{The assumption that $\R(X)$ and $\V$ are finite is also
necessary
for the abbreviation $X \leadsto Y$ used in C6 to be in $\Lprop$; however,
we can replace C6 by the axiom scheme
$$\neg(\land_{i=0}^{k-1}(X_{i+1})_{\vec{y}_ix_i}(\vec{u}_i) = z_i \land
(X_{i+1})_{\vec{y}_i} = z_i') \land
(X_{0})_{\vec{y}_kx_k}(\vec{u}_k) = z_k \land
(X_{0})_{\vec{y}_k} = z_k'),$$  where $x_i \in \R(X_i)$ for $i = 1,
\ldots, k$.  That is, we essentially replace C6 by
all its instances.  This axiom is equivalent to C6 (although not as
transparent) and can be expressed even if $|\V|$ is infinite or
$|\R(X)|$
is infinite for some $X \in \V$.}
C3--C5 were introduced by Galles and Pearl
\citeyear{GallesPearl97,GallesPearl98}, as were their names.  Roughly
speaking, C3
says that if the value of $W$ is $w$ in all solutions to the equations
$T_{\vec{x}}(\vec{u})$, then all solutions to the equations in
$T_{\vec{x}w}(\vec{u})$ are the same as the solutions to the
equations in $T_{\vec{x}}(\vec{u})$.  C3 is valid in $\T$ as well as
$\Tun$.  As we shall see, a variant of C3 (obtained by replacing ``all''
by ``some'') is also valid in $\T$.
C4 simply says
that in all solutions obtained after setting $X$ to $x$, the value of
$X$ is $x$.
C5 is perhaps the least obvious of these
axioms; the proof of its soundness is not at all straightforward. It
says that
if setting $\vec{X}$ to $\vec{x}$ and $W$ to $w$ results in $Y$ having
value $y$ and setting $\vec{X}$ to $\vec{x}$ and $Y$ to $y$ results in
$W$ having value $w$, then $Y$ must already have value when we set
$\vec{X}$ to $x$ (and $W$ must already have value $w$).

Finally, it is easy to see that C6 holds in recursive models.
For if $Y \leadsto Z$, then $Y$ must precede $Z$ in the
causal ordering.  Thus, if $X_0 \leadsto X_1 \land \ldots \land X_{k-1}
\leadsto X_k$, then $X_0$ must precede $X_k$ in the causal ordering,
so $X_k$ cannot affect $X_0$.  Thus, $\neg(X_k \leadsto X_0)$ holds.
As we shall see, in a precise sense, C6 characterizes recursive models.

C6 can be viewed as a collection of axioms (actually, axiom schemes),
one for each $k$.  The case $k=1$ already gives us
$\neg(Y \leadsto Z) \lor \neg(Z
\leadsto Y)$ for all variables $Y$ and $Z$.  That is, it tells us
that, for any pair of variables, at most one affects the other.
However, just restricting C6 to the case of $k=1$ does not suffice to
characterize $\Trec$, as the following example shows.

\xam \label{C6notsuff}
Let $\S = (\emptyset, \{X_0,X_1,X_2\}, \R)$,
where $\R(X_0) = \R(X_1) = \R(X_2) = \{0,1,2\}$, and let
$T = (\S, \F)$, where
$F_{X_i}$ is characterized by the equation
$$
X_i = \left \{
\begin{array}{ll}
2 &\mbox{if $X_{i \oplus 1} = 1$}\\
0 &\mbox{otherwise}
\end{array} \right.
$$
and $\oplus$ is addition mod 3.
It is easy to see that $T \in \Tun$:  If any of the variables are
set, the equations completely determine the values of all the other
variables.  On the other hand, if none of the variables are set, it is
easy to see that $(0, 0, 0)$ is the only solution that satisfies
all the equations.  Moreover, in $T_{\vec{X} \gets \vec{x}}$, the
variable $X_i$ is 0 unless it is set to a value other than 0 or $X_{i
\oplus 1}$ is set to 1.  It easily follows that
$X_i$ is affected only by $X_{i \oplus 1}$.  A straightforward
verification (or an appeal to
Theorem~\ref{thm1} below) shows that $T$ satisfies all the axioms other
than
C6. C6 does not hold in $T$, since $T \sat X_0 \leadsto X_1 \land X_1
\leadsto X_2 \land X_2 \leadsto X_0$.  This also shows that $T$ is not
recursive.  However, the restricted version
of C6 (where $k=1$) does hold in $T$.  A generalization of this example
(with $k$ random variables rather than just 2) can be used to show that
we cannot bound $k$ at all in C6; we need C6 to hold for all finite
values of $k$.
\exam

Let $\AXun(\S)$ consist of C0--C5 and MP;
let $\AXrec(\S)$ consist of C0--C4, C6, and MP.
We could include C5 in $\AXrec(\S)$; I did not do so because, as Galles
and Pearl \citeyear{GallesPearl98} point out, it follows from C3 and
C6.  Note that the signature $\S$ is a parameter of the axiom system,
just as it is for the language and the set of models.  This is because,
for example, the set $\R(X)$ (which is determined by $\S$) appears
explicitly in C1 and C2.

\thm\label{thm1}
$\AXun(\S)$ (resp., $\AXrec(\S)$) is a sound and complete
axiomatization for $\Lprop(\S)$ with respect to $\Tun(\S)$ (resp.,
$\Trec(\S)$).
\ethm

\prf  See the appendix.  \eprf

As I said in the introduction, Galles and Pearl \citeyear{GallesPearl98}
prove a similar
completeness result for causal models whose variables
satisfy a fixed causal ordering.  Given a total ordering
$\prec$ on the variables in $\V$, consider the following axiom:
\begin{itemize}
\item[Ord.] $Y_{\vec{x}w}(\vec{u}) = Y_{\vec{x}}(\vec{u})$ if $Y \prec W$
\end{itemize}
Since $\vec{x}$, $w$, and $\vec{u}$ are implicitly universally
quantified in Ord, this axiom says that $\neg (W
\leadsto Y)$ holds if $Y \prec W$.  It follows that if $W \leadsto Y$,
then $W \prec Y$.  From this and the fact that $\prec$ is a total order,
it is easy to see that Ord implies C6.

Galles and
Pearl show that C1--C4 and Ord
is a sound and complete axiomatization with respect to the class of
causal models satisfying Ord for $\LGP$.
More precisely, Galles and Pearl take $A_C$ to
consist of the axioms C1--C4 and Ord (but not C0 or MP),
and show, in their notation, that
$S \sat \sigma$ implies $S \stur_{A_C} \sigma$, where $S \union
\{\sigma\}$ is a set of formulas in $\LGP$.
There is an important subtle point worth stressing about their result:
C1 and C2, which are axioms in $A_C$, are not expressible in $\LGP$
(since their statement involves disjunction and negation).

So what exactly is Galles and Pearl's result saying?
They interpret $S
\sat \sigma$, as usual, as meaning that in all causal models
satisfying $S$, $\sigma$ is true.%
\footnote{Although they do not say this
explicitly, it is clear that they intend to further restrict to
casual models satisfying $S$ {\em and Ord}, for the fixed order $\prec$.
Without this restriction, their result is not true.}
They interpret $S \stur_{A_C} \sigma$ as meaning that $\sigma$ is
provable from $S$ and the axioms in the axioms of $A_C$ ``together with
the rules of logic'', which presumably means C0 and MP.  It follows
easily from Theorem~\ref{thm1} that their result is correct (see below),
but it is unlike typical soundness and completeness
proofs, since the proof of $\sigma$ from $S$ will in general involve
formulas in $\Lprop$ that are not in $\LGP$.  (In particular,
this will happen whenever C1--C3 are used in the proof.)

To see that Galles and Pearl's result follows from Theorem~\ref{thm1},
define $S^*$ to be
the formula in $\Lprop(\S)$ which is the conjunction of the formulas in
$S$
(there can only be finitely many, since $\LGP(\S)$ itself has only
finitely
many distinct formulas), together with the conjunction of all the
instances of the axiom Ord (again, there are only finitely many).
Note that $S \sat \sigma$ holds iff $\Tun(\S) \sat S^* \rimp \sigma$
(since the formulas in Ord guarantee that the only causal models
that satisfy $S^*$ are recursive, and hence are in $\Tun(\S)$).  Thus,
by Theorem~\ref{thm1}, $S \sat \sigma$ iff $\AXun(\S) \vdash S^* \rimp
\sigma$.
The latter statement is equivalent to $S \vdash_{A_C} \sigma$, as
defined by Galles and Pearl.  In fact, Theorem~\ref{thm1} shows that
$\AXun(\S) + \mbox{Ord}$ gives a sound and complete axiomatization with
respect to causal models satisfying Ord for the language $\Lprop(\S)$,
which allows Boolean connectives.  (Of course, Theorem~\ref{thm1} shows
more, since it extends Galles and Pearl's result to $\Trec(\S)$ and
$\Tun(\S)$.)
This suggests that $\Lprop$ is a more appropriate language for reasoning
about causality than $\LGP$, at least for causal models in $\Tun$.
$\LGP$ cannot express a number of properties of
causal reasoning of interest (for example, the ones captured by axioms
C1--C3).
When we use $\Lprop$, not only is every formula in $\Lprop$ valid in
$\Tun$ provable from the axioms in $\AXun$, but the proof involves only
formulas in $\Lprop$.

What about $\T$?  I have not been able to find a complete axiomatization
for the language $\Lprop$ with respect to $\T$.  However, I do not think
that finding a complete axiomatization for $\Lprop$ with respect to $\T$
is of great interest, because
$\Lprop$ is simply not a language appropriate for reasoning about
causality in $\T$.  Because there is not necessarily a unique
solution to the equations that arise in a causal model $T \in \T$,
it is useful to be able to say both that there exists a solution
with certain properties and that {\em all\/} solutions have certain
properties.  This is precisely what the language $\Lex$ lets us do.%
\footnote{Note that $\Lex$ allows us to
say
that there is a
unique solution for a random variable $X$ after setting some other
variables.   For example, $\<\vec{Y} \gets
\vec{y}\> \true(\vec{u}) \land  [\vec{Y} \gets \vec{y}](X(\vec{u}) = x)$
says that there are solutions to the equations when $\vec{Y}$ is set to
$\vec{y}$ and $\U$ is set to $\vec{u}$ and,
in all of them, $X$ is uniquely determined to be $x$.}
 As I now show, there is in fact an elegant sound
and complete axiomatization for $\Lex$ with respect to $\T$.

Consider the following axioms:
\begin{itemize}
\item[D0.] All instances of propositional tautologies.
\item[D1.] $[\vec{Y} \gets \vec{y}](X(\vec{u}) = x \rimp
X(\vec{u}) \ne x')$  if $x, x' \in \R(X)$, $x \ne x'$ \hfill
(functionality)
\item[D2.] $[\vec{Y} \gets \vec{y}](\lor_{x \in \R(X)} X(\vec{u}) = x)$
\hfill (definiteness)
\item[D3.] $\<\vec{X} \gets \vec{x}\>(W(\vec{u}) = w
 \land \vec{Y}(\vec{u}) = \vec{y})
  \rimp \<\vec{X} \gets \vec{x};W \gets
w\>(\vec{Y}(\vec{u}) = \vec{y})$
\hfill
(composition)
\item[D4.] $[\vec{W} \gets \vec{w}; X \gets x](X(\vec{u}) = x)$ \hfill
(effectiveness)
\item[D5.] $(\<\vec{X}\gets \vec{x}; Y \gets y\> (W(\vec{u}) = w \land
\vec{Z}(\vec{u}) = \vec{z})  \land
\<\vec{X}\gets \vec{x}; W \gets w\> (Y(\vec{u}) = y \land
\vec{Z}(\vec{u}) = \vec{z}))$\\
$\mbox{ }\ \ \ \rimp \<\vec{X} \gets \vec{x}\> W(\vec{u})
= w \land Y(\vec{u}) = y \land \vec{Z}(\vec{u}) = \vec{z}))
$, where $\vec{Z} = \V - (\vec{X} \union \{W,Y\})$\\
\mbox{ } \hfill
(reversibility)
\item[D6.] $(X_0 \leadsto X_1 \land \ldots \land X_{k-1} \leadsto X_k)
\rimp
\neg (X_k \leadsto X_0)$ \hfill (recursiveness)
\item[D7.] $([\vec{X} \gets \vec{x}]\phi \land [\vec{X} \gets
\vec{x}](\phi \rimp \psi)) \rimp  [\vec{X} \gets \vec{x}]\psi$
 \hfill
(distribution)
\item[D8.] $[\vec{X} \gets \vec{x}]\phi$ if $\phi$ is a propositional
tautology  \hfill (generalization)
\item[D9.] $\langle \vec{Y} \gets \vec{y} \rangle \true(\vec{u}) \land
\lor_{x
\in \R(X)} [\vec{Y} \gets {\vec{y}}](X(\vec{u}) = x)$ if $Y = \V -
\{X\}$\\
\mbox{ } \hfill (unique
solutions for $\V - \{X\}$)
\item[D10.] $\langle \vec{Y} \gets \vec{y} \rangle \true(\vec{u}) \land
\lor_{x
\in \R(X)} [\vec{Y} \gets {\vec{y}}](X(\vec{u}) = x)$ \mbox{ } \hfill
(unique solutions)
\item[D11.] $\langle \vec{Y} \gets \vec{y} \rangle (\phi_1(\vec{u}_1)
\land \ldots \land \phi_k(\vec{u}_k)) \dimp
(\langle \vec{Y} \gets \vec{y} \rangle \phi_1(\vec{u}_1)
\land \ldots \land
\langle \vec{Y} \gets \vec{y} \rangle \phi_k(\vec{u}_k)$, if
$\phi_i(\vec{u}_i)$ is a Boolean combination of formulas of the form
$X(\vec{u}_i) = x$ and
$\vec{u}_i \ne \vec{u}_j$ for $i \ne j$
\hfill (separability)
\end{itemize}

D1--D6 are the analogues of C1--C6 in $\Lex$.  D4 and D6 are just C4 and
C6, with no changes at all.  The other axioms are not quite the same
though.  For example,
C1 is actually $[\vec{Y} \gets \vec{y}](X(\vec{u}) = x) \rimp
\neg [\vec{Y} \gets \vec{y}] (X(\vec{u}) = x')$  if $x \ne x'$.
By Lemma~\ref{uniquelem}, this is equivalent to D1 in $\Tun$; however,
the two formulas are not equivalent in general.  Similarly, C2 is
$\lor_{x \in \R(X)} [\vec{Y} \gets \vec{y}](X(\vec{u}) = x)$, which is
closer to D10 than D2 (since the disjunction is outside the scope of the
$[\vec{Y}\gets \vec{y}]$).   Again, D10 and D2 are equivalent in
$\Tun$ (both are equivalent to C2 in this case)
but, in general, D10 is stronger than D2.  Only D2 and D9, both weaker
than D10, hold in $\T$.  The exact analogue of C3 would use $[\,]$
instead of $\< \, \>$ and
say $Y(\vec{u}) = y$ instead of $\vec{Y}(\vec{u}) = \vec{y}$.
For completeness, it is necessary to have a vector of variables
here.  Using $[
\,]$ instead of $\< \, \>$ also results in a valid formula (and would
not require a vector $\vec{Y}$).  While the
two variants are equivalent in $\Tun$, they are different in general,
and the one given here is the more useful.  (More precisely, with it we
get completeness, while the version with $[ \,]$ does not suffice for
completeness.) Similarly, in D5, we use
$\<\,\>$ instead of $[ \, ]$, and add the extra clause $\vec{Z}(\vec{u})
= \vec{z}$.  Both turn out to be necessary for soundness.
In some sense, we can think of D1--D6 as capturing
the ``true content'' of C1--C6, once we drop the assumption that the
structural equations have a unique solution.  D7 and D8 are standard
properties of modal operators.  D10 is what we need to capture the fact
that the structural equations have unique solutions.  D11
essentially says that
the solutions to the equations that arise  when the exogenous variables
are set to
$\vec{u}$ are independent of the solutions that arise when the exogenous
variables are set to $\vec{u}' \ne \vec{u}$.

Let $\AXex$ consist of D0--D5, D7--D9, D11, and  MP (modus
ponens); let $\AXunex$ be the result of adding D10 to $\AXex$; let
$\AXrecex$ be the result of adding D6 to $\AXunex$.

\thm\label{thm2}
$\AXex(\S)$ (resp., $\AXunex(\S)$, $\AXrecex(\S)$) is a sound and
complete axiomatization for $\Lex(\S)$ with respect to $\T(\S)$ (resp.,
$\Tun(\S)$,
$\Trec(\S)$).
\ethm

\prf  See the appendix.  \eprf

\section{Decision Procedures}\label{decisionp}
In this section I consider the complexity of deciding if a formula is
satisfiable (or valid).  This, of course, depends on the language
($\Lex$, $\Lprop$, or $\LGP$) and the class of models ($\Trec$, $\Tun$,
$\T$) we consider.  It also depends on how we formulate the problem.

One version of the problem is to consider a fixed signature $\S =
(\U,\V,\R)$, and
ask how hard it is to decide if a formula $\phi \in \Lex(\S)$ (\respc
$\Lprop(\S)$, $\LGP(\S)$) is satisfiable in $\Trec(\S)$ (\respc
$\Tun(\S)$, $\T(\S)$).  If $\S$ is finite
(that is, if $\V$ and $\U$ are finite and $\R(Y)$ is
finite for each $Y \in \U \in \V$), this turns out to be quite easy, for
trivial reasons.

\thm\label{decprob1} If $\S$ is a fixed finite signature, the problem of
deciding if a formula $\phi \in \Lex(\S)$ (\respc $\Lprop(\S)$,
$\LGP(\S)$) is satisfiable in $\Trec(\S)$ (\respc $\Tun(\S)$, $\T(\S)$)
can be solved in time linear in $|\phi|$
(the length of $\phi$ viewed as a string of symbols). \ethm

\prf If $\S$
is finite, there are only finitely many causal models in $\T(\S)$,
independent of $\phi$.
Given $\phi$, we can explicitly check if $\phi$ is satisfied in any (or
all) of them.  This can be done in time linear in $|\phi|$.  Since $\S$
is not a parameter to the problem, the huge number of possible
causal models that we have to check affects only the constant. \eprf

We can do even better than Theorem~\ref{decprob1} suggests if $\S$ is a
fixed finite signature.  Suppose that $\V$ consists of 100 variables and
$\phi$ mentions only 3 of them.  A causal model must specify the
equations for all 100 variables.   Is it really necessary to consider
what happens to the 97 variables not mentioned in $\phi$ to decide if
$\phi$ is satisfiable or valid?  As the following result shows, if we
restrict to models in $\Tun$, then we need to check only the variables
that appear in $\S$.
Given a signature $\S = (\U,\V,\R)$, let
$\S_\phi = (\{U^*\},\V_\phi,\R_\phi)$,
where $\V_\phi$ consists of the variables in $\V$ that appear in $\phi$,
$U^*$ is a fresh exogenous variable, not mentioned in $\V$ or $\U$,
$\R_\phi(X) = \R(X)$ for $X \in \V_\phi$, and $\R_\phi(U^*)$ consists of
all
those tuples in $\times_{U \in \U} \R(U)$ that are mentioned in $\phi$.

\thm\label{finite1} A formula $\phi \in \Lex(\S)$ is satisfiable in
$\Trec(\S)$ (\respc $\Tun(\S)$) iff it is satisfiable in
$\Trec(\S_\phi)$ (\respc $\Tun(\S_\phi)$). \ethm

\prf See the appendix.  \eprf

The analogue to Theorem~\ref{finite1} does not hold for $\T$.  For
example, suppose that $\S =(\emptyset,\{X,Y,Z\},\R)$, where
$\R(X) = \R(Y) = \R(Z)= \{0,1\}$, and
$\phi$ is the formula  $\<X \gets 0\>(Y = 0) \land \<X \gets
0\>(Y = 1)$.  It is easy to see that there is a causal model
in $\T(\S)$ satisfying $\phi$.  For example, if
$T=(\S,\F)$, where $F_X(y,z) = y \oplus z$, $F_Y(x,z) = x
\oplus z$ and $F_Z(x,y) = x \oplus y$, and $\oplus$ represents addition
mod 2, then it is easy to check that $T \sat \phi$.  On the other hand,
there is no causal model $T' \in \T(\S_\phi)$ such that $T' \sat
\phi$.  For suppose $T' \sat \phi$ and $T' = (\S_{\phi},\F')$.
Since $T' \sat \<X \gets 0\>(Y = 0)$, we must have $F_Y'(0) =
0$; since $T' \sat \<X \gets 0\>(Y = 1)$, we must have $F_Y'(0) = 1$.
But we cannot have both $F_Y'(0) = 0$ and $F_Y'(1) = 1$, since $F_Y'$ is
a function.

There is a variant of Theorem~\ref{finite1} that does hold for $\T$
that does give us a bound on the number of variables we need to
consider.
Given a signature $\S = (\U,\V,\R)$,
define $||\S|| = \times_{X \in \V} |\R(X)|$ (where we take $||S|| = \infty$
if either $\V$ is infinite or $|\R(X)| = \infty$ for some $X \in \V$).
If $||\S|| > ||\S_\phi||^2 + ||\S_\phi||$, let
$\S_\phi^+ =
(\{U^*\},\V_\phi^+,\R_\phi^+)$,
where $\V_\phi^+$ is $\V_\phi$ as defined above together with
one fresh
endogenous variable $X^*$, $\R_\phi^+(X^*) = \times_{X \in
\V_\phi} \R(X)$, and $\R_\phi^+(U^*) = \R_\phi(U^*)$.
If $||\S|| \le ||\S_\phi||^2 + ||\S_\phi||$, let
$\S_\phi^+ =
(\{U^*\},\V,\R')$, where $\R'(X) = \R(X)$ for $X \in \V$ and $\R'(U^*) =
\R_\phi(U^*)$.

\thm\label{finite1a}
A formula $\phi \in \Lex(\S)$ is satisfiable in
$\T(\S)$ iff $\phi$ is satisfiable in $\T(\S_\phi^+)$.  \ethm

\prf See the appendix.  \eprf

Note that if $||\S|| \le ||\S_\phi||^2 + ||\S_\phi||$, then, since we
have assumed
(without loss of generality) that $|\R(X)| \ge 2$ for each variable $X$,
it must be the case that there are at most $2 \log_2(||\S_\phi||) + 1$
variables in signature $\S$.

Since Theorems~\ref{finite1} and~\ref{finite1a} apply to all formulas
in $\Lex(\S)$, they apply
{\em a fortiori\/} to formulas in $\Lprop(\S)$ and $\LGP(\S)$.
Although stated only in terms of satisfiability, it is immediate that
they also hold for validity.  Thus, they tell us that, without loss of
generality, when considering satisfiability or validity, we need to
consider only finitely many variables (essentially, the ones that
appear in $\phi$, and perhaps a few more).
In this sense, we can restrict to signatures with only finitely many
variables without loss of generality.  Note that these results do {\em
not\/} tell us that we can restrict to finite sets of values for these
variables without loss of generality.

Returning to the complexity of the decision problem,
note that Theorem~\ref{decprob1} is the analogue of the observation that
for propositional logic, the satisfiability problem is in linear time if
we restrict to a fixed set of primitive propositions.  The proof that
the satisfiability problem for propositional logic is NP-complete
implicitly assumes that we have an unbounded number of primitive
propositions at our disposal.

There are two ways to get an analogous result here.  The first is to
allow the signature $\S$ to be infinite and the second is to make the
signature part of the input to the problem.  The results in both
cases are similar, so I just consider the case where the signature is
part of the input here.

\thm\label{decprob2}  Given as input a pair $(\phi,\S)$, where $\phi
\in \Lex(\S)$ (\respc $\Lprop(\S)$) and $\S$ is a finite
signature, the problem of deciding
if $\phi$ is satisfiable in $\Trec(\S)$ (\respc $\Tun(\S)$,
$\T(\S)$) is NP-complete (\respc NP-hard) in $|\phi|$;
if $\phi \in \LGP(\S)$, then the problem of deciding if $\phi$ is
satisfiable in $\Trec(\S)$ (\respc $\Tun(\S)$) is NP-complete (\respc
NP-hard).
\ethm

\prf See the appendix. \eprf

I believe that the problem of deciding if a formula $\phi$ in
$\Lprop(\S)$ or $\Lex(\S)$ is satisfiable in $\Tun(\S)$ and
$\T(\S)$ is NP-complete, as is the case of deciding if $\phi \in
\LGP(\S)$ is satisfiable in $\Tun(\S)$.
However, I have not been able to show this.
What about the satisfiability problem for formulas in $\LGP$ in
$\T(\S)$?  This may well be in constant time!  Indeed, if $\S$ is an
infinite signature (that is, if $\S = (\U,\V,\R)$ and $|\V| = \infty$),
then it is provably in constant time.  The point is that
a formula in $\LGP(\S)$ is trivially satisfiable in a structure $T \in
\LGP(\S)$ where for all settings $\vec{X} \gets \vec{x}$, the equations
in $T_{\vec{X} \gets \vec{x}}$ have no
solutions, and there always is such
model structure if $\S$ has infinitely many variables.  If $\S$ has only
finitely many variables, we do not have such trivial models, but it may
still be possible to show that a ``trivial enough'' model exists that
satisfies the formula.  This just emphasizes that $\LGP(\S)$ is simply
too weak a language to reason about models in $\T(\S)$.

\commentout{
 Results similar to
Theorem~\ref{decprob2}
also hold for the case where $\S$ is a fixed infinite signature, with no
change in proof.

\thm\label{decprob4} If $\S$ is a fixed infinite signature, the problem
of deciding if a formula $\phi \in \Lex(\S)$ (\respc $\Lprop(\S)$
is
satisfiable in $\Trec(\S)$ (\respc $\Tun(\S)$, $\T(\S)$) is NP-complete
(\respc NP-hard) in $|\phi|$.
\ethm

Notice that the language $\LGP$ is not mentioned in
Theorems~\ref{decprob2} and~\ref{decprob4}.  This is because the
situation for $\LGP$ is different from the other languages, and harder
to characterize.  The following theorem gives some sense of the types of
results that can be proved.  Among other things, it shows
that if we have a fixed signature,
the complexity of the satisfiability problem depends on the cardinality
of the sets $\R(X)$ for $X \in \V$.  Roughly speaking, it becomes
easier to check if a formula is satisfiable if the cardinality of
these sets is {\em large}.  For example, if $|\R(X)| = 2$ for all $X \in
\V$, then a slight variant of the proof of
Theorem~\ref{decprob2} shows satisfiability problem is NP-hard, just
as before. However, if $|\R(X)| = \infty$ for all $X \in \V$, then the
satisfiability problem is polynomial time!

\thm\label{decprob5}
\begin{itemize}
\item[(a)] Given as input a pair $(\phi,\S)$, where $\phi
\in \LGP(\S)$ and $\S$ is a finite or infinite
signature, the problem of deciding
if $\phi$ is satisfiable in $\Trec(\S)$
(\respc $\Tun(\S)) is NP-complete (\respc NP-hard).
\item[(b)]
Suppose $\S = (\V,\R)$ is a fixed infinite
signature, where $\V = \{X_1, X_2, X_3, \ldots\}$
and $f(k) = |\V_k|$.
\begin{itemize}
\item[(i)] The problem
of deciding if a formula $\phi \in \LGP(\S)$ is
satisfiable in $\Trec(\S)$ is NP-complete (in $|\phi|$).
\item[(ii)] If there is some polynomial $p$ such that $f(k) \le p(k)$
for all $k$ and $f$ itself is polynomial time computable,
then the problem of deciding if a formula $\phi \in \LGP(\S)$
is satisfiable in $\Tun(\S)$ is NP-hard (in $|\phi|$).
\item[(iii)] If $f(k) = \infty$ for all
$k$, then the problem of deciding if a formula $\phi \in \LGP(\S)$ is
satisfiable in $\Tun(\S)$ is in polynomial time (in
$|\phi|$).  Morever, the problem of deciding if $\phi_1 \rimp \phi_2$ is
valid for $\phi_1, \phi_2 \in \LGP(\S)$ is polynomial time.%
\footnote{In the conference version of this paper (in {\em Proceedings
of the Fourteenth Conference on Uncertainty in Artificial Intelligence},
1998, pp.~202--210), I claimed that if $\S$ is infinite and $|\R(X)|=
\infty$ for all but finitely many variables in $\S$, then the problem
of deciding if $\phi \in \LGP(\S)$ is satisfiable in $\Trec(\S)$ is in
polynomial
time. As Theorem~\ref{decprob5} shows, this is not correct
for $\Trec(\S)$, but is correct for $\Tun(\S)$ and $\T(\S)$.}
\item[(iv)] The problem
of deciding if a formula $\phi \in \LGP(\S)$ is
satisfiable in $\T(\S)$ is in constant time.
\end{itemize}
\end{itemize}
\ethm

\prf See the appendix. \eprf

The proof of part (a) is very similar to that of Theorem~\ref{decprob2}.
New subtleties arise in part (b).
The NP lower bound in part~(i) is due
(in part) to the uncertainty about the ordering of the variables.  If we
assume a fixed ordering $\prec$ on the variables (as Galles and Pearl
\citeyear{GallesPearl98} do) and restrict to causal models in
$\Trec(\S)$ where the ordering of the variables satisfies $\prec$, then
the analogue of parts (ii) and (iii) holds.
The bounds given in (ii) and (iii) of part
(b) are in no way tight. For example,
to get the NP lower bound of (ii), it suffices that, there is a
polynomially computable function $g$ such that $|\R(X_{g(k)}| \le
p(k)$. In fact, (ii) is the special case where $g$ is the identity;
essentially the same proof works for all $g$.
Part (iii) also works if $f(k) =
2^{2^k}$, using essentially the same proof.  Thus, we can get
polynomial time results even without required infinite domains for the
variables.
Finally, it is worth noting that the NP lower bound in part~(i) is due
(in part) to the uncertainty about the ordering of the variables.  If we
assume a fixed ordering $\prec$ on the variables (as Galles and Pearl
\citeyear{GallesPearl98} do) and restrict to causal models in
$\Trec(\S)$ where the ordering of the variables satisfies $\prec$, then
the analogue of parts (ii) and (iii) holds.  The constant time result of
part (iv) follows from the observation that if $\S$ is infinite, there
is always a causal model in $\T(\S)$ where there are no solutions to any
equations, so any formula $\phi \in \LGP(\S)$ is trivially satisfied.
}%

\section{Conclusion}\label{conclusion}
I have provided complete axiomatizations and decision procedures
for propositional languages for
reasoning about causality.  I have tried to stress the important role of
the choice of language (and the signature)
in both the axiomatizations and, more generally,
in the reasoning process.

Both the models and the languages considered here are somewhat limited.
For example, a more general approach to modeling causality would allow
there to be more than one value of $X$ once we have set all the other
variables.  This would be appropriate if we model things at a somewhat
coarser level of granularity, where the values of all the variables
other than $X$ do not suffice to completely determine the value of $X$.
I believe the results of this paper can be extended in a
straightforward way to deal with this generalization, although I have
not checked the details.  For general causal reasoning, I
believe we need a richer language, which includes some first-order
features.  I hope to return to the issue of finding appropriate richer
languages for causal reasoning in future work.

\acks I'd like to thank Judea Pearl for his
comments on a previous version of this paper, as well as his generous
help in providing pointers to the literature.
This work was supported in part by NSF under
grant IRI-96-25901 and by the Air Force Office of
Scientific Research under grant F49620-96-1-0323.
A preliminary version of this
paper appears in {\em Proc.~Fourteenth Conference on
Uncertainty in AI}, pp.~202--210, 1998.

\appendix
\section{Proofs}

\othm{thm1}
$\AXun$ (resp., $\AXrec$) is a sound and complete
axiomatization for $\Lprop(\S)$ with respect to $\Tun(\S)$ (resp.,
$\Trec(\S)$).
\eothm

\medskip

\prf Soundness is proved by Galles and Pearl.  To make the paper
self-contained, I reprove the only non-obvious case---the validity of
C5 in $\Tun$.

Let $T \in \Tun$ and suppose that $T \sat Y_{\vec{x}w}(\vec{u}) =
y \land W_{\vec{x}y}(\vec{u}) = w$.
We want to show that $T \sat Y_{\vec{x}}(\vec{u}) = y$.
Since we are in $\Tun$, there
is a unique vector $\vec{v}_1$ that satisfies the equations in
$T_{\vec{x}w}(\vec{u})$ and a unique vector $\vec{v}_2$ that satisfies
the equations in $T_{\vec{x}y}(\vec{u})$.  I claim that $\vec{v}_1 =
\vec{v}_2$.
By assumption, the $\vec{X}$, $Y$, and $W$ components of these vectors
are the same ($\vec{x}$, $y$, and $w$, respectively).
Now consider the $T_{\vec{x}yw}(\vec{u})$.  I
claim that
$\vec{v}_1$ and $\vec{v}_2$ are both solutions to the equations in that
causal theory.  Note that for any variable $Z$ other than those in
$\vec{X}
\union \{W,Y\}$, the equation $F_Z^{\vec{x}w,\vec{u}}$ for $Z$ in
$T_{\vec{x}w}(\vec{u})$ is the same as the equations
$F_Z^{\vec{x}y,\vec{u}}$ and $F_Z^{\vec{x}yw,\vec{u}}$
for $Z$ in $T_{\vec{x}y}(\vec{u})$ and $T_{\vec{x}yw}(\vec{u})$,
respectively, except that in
the first case, $w$ has been plugged in as the value of $W$, in the
second case $y$ has been plugged in as the value of $Y$, and in the
third case, both $w$ and $y$ have been plugged in. However, since
$w$ and $y$ are the values of $W$ and $Y$, respectively, in both
$\vec{v}_1$ and $\vec{v}_2$, and since these vectors satisfy both
equation $F_Z^{\vec{x}w}$ and $F_Z^{\vec{x}y}$,
they must
also satisfy $F_Z^{\vec{x}wy}$.  Since the
equations for $T_{\vec{x}yw}(\vec{u})$ have a unique solution,
we have that $\vec{v}_1 = \vec{v}_2$, as desired.

Next, I claim that $\vec{v}_1$ satisfies the equations in $T_{\vec{x}}
(\vec{u})$.
Again, as above, it is clear that it satisfies the equation for $Z
\notin \vec{X} \union \{W,Y\}$.  A similar argument shows that it
satisfies the equation for $Y$ in $T_{\vec{x}}
(\vec{u})$, since $\vec{v}_1$
satisfies the equation for $Y$ in $T_{\vec{x}w}(\vec{u})$.
Finally, a similar argument shows that it satisfies the equation for
$W$ in $T_{\vec{x}}(\vec{u})$, since $\vec{v}_2 =
\vec{v}_1$ satisfies the equation for $W$ in $T_{\vec{x}y}(\vec{u})$.
Since the $Y$ component of
$\vec{v}_1$ is $y$, it follows that $Y_{\vec{x}}(\vec{u}) = y$.

So much for soundness.
For completeness, as usual, it suffices to prove that if a formula in
$\Lprop$ is consistent with $\AXun$ (resp., $\AXrec$), then it is
satisfied in a causal model in $\Tun$ (resp., $\Trec$).  (Here's the
argument:  We want to show that every valid formula is provable.
Suppose that we have shown that every consistent formula is
satisfiable and that $\phi$ is valid.  If $\phi$ is not provable, then
$\neg \phi$ is consistent.  By assumption, this means that $\neg \phi$
is satisfiable, contradicting the assumption that $\phi$ is valid.)

I now give the argument in the case of $\AXun$.

Suppose that a formula $\phi \in \Lprop(\S)$, with $\S =
(\U,\V,V)$, is consistent with $\AXun$.
Consider a maximal consistent set $C$ of formulas that includes $\phi$.
(A maximal
consistent set is a set of formulas whose conjunction is consistent such
that any larger set of formulas would be inconsistent.)
It follows easily
from standard propositional reasoning (i.e., using C0 and MP only) that
such a maximal consistent set exists.  Moreover, from C1 and C2, it
follows that for each random variable $X \in \V$ and vector $\vec{y}$ of
values,
there exists exactly one element $x \in \R(X)$ such that $X_{\vec{y}} =
x \in C$.  I now construct a causal model $T = (\S,F) \in \Tun(\S)$
that
satisfies every formula in $C$ (and, in particular, satisfies $\phi$).

A term $X_{\vec{Y} \gets \vec{y}}(\vec{u})$ is {\em complete (for
$X$)\/} if
$\vec{Y}$ consists of all the variables in $\V - \{X\}$.
Thus, $X_{\vec{Y} \gets \vec{y}}(\vec{u})$ is
a complete term if every random variable other than $X$ is determined.
We use the complete terms to define the structural equations.
For each variable in $X\in \V$,
define $F_X(\vec{u},\vec{y}) = x$ if $X_{\vec{y}}(\vec{u}) = x$, where
$X_{\vec{y}}(\vec{u})$ is a complete term.
This gives us a causal model $T$.  Now we have to show that
this model is in $\Tun$ and that all the formulas in $C$ are satisfied
by $T$.

I show that $X_{\vec{Y} \gets \vec{y}}(\vec{u}) = x$ is in $C$ iff $T
\models X_{\vec{Y} \gets \vec{y}}(\vec{u}) = x$
by induction on $|\V| - |\vec{Y}|$.  The case where $|\V| - |\vec{Y}| =
0$ follows immediately from C4, since then $X$ is in $\vec{Y}$.  If
$|\V| - |\vec{Y}| \ne 0$, we can assume without loss of generality that
$X$ is not in $\vec{Y}$, for otherwise the result again follows from C4.
Given this assumption, if $|\V| - |\vec{Y}| = 1$, the result follows by
definition of the equations $F_X$.

For the general case, suppose that $|\V| - |\vec{Y}| = k >1$. We want
to show that there is a unique solution to the equations in $T_{\vec{Y}
\gets \vec{y}}(\vec{u})$ and that, in this solution, $X$ has value $x$.
To see that there is a solution, we define a vector $\vec{v}$ and show
that it is in fact a solution.
If $W \in \vec{Y}$ and
$W\gets w$ is the assignment to $W$ in $\vec{Y} \gets \vec{y}$,
then we set the $W$ component of $\vec{v}$ to $w$.  If $W$ is not
in $\vec{Y}$, then set the
$W$ component of $\vec{v}$ to the unique value
$w^*$ such that $W_{\vec{Y} \gets \vec{y}}(\vec{u}) = w^*$ is in $C$.
(By C1 and
C2 there is such a unique value $w$.)  I claim that $\vec{v}$ is a
solution to the equations in $T_{\vec{Y} \gets \vec{y}}(\vec{u})$.

To see this, let $W$ be a variable in $\V$ not in $\vec{Y}$.
Let $\vec{Y}' = \vec{Y}W$.  By C3 and C4, for every
variable $Z \in \V - \vec{Y}'$, we have
$Z_{\vec{y}w^*}(\vec{u}) = z^*$.  Since $|\V| - |\vec{Y}'| = k-1$, by the
inductive hypothesis, $\vec{v}$ is in fact
the unique solution for $T_{\vec{y}w^*}(\vec{u})$.  For every variable
$Z$ in $\V - \vec{Y}'$, the equation $F_Z^{\vec{y}w^*,\vec{u}}$ for $Z$
in $T_{\vec{y}w^*}(\vec{u})$ is the same as
the equation $F_Z^{\vec{y},\vec{u}}$
for $Z$ in $T_{\vec{y}}(\vec{u})$, except that $W$ is set to $w^*$.
Thus,  every equation in $T_{\vec{y}}(\vec{u})$ except possibly the
equation $F_W^{\vec{y},\vec{u}}$ is satisfied by $\vec{v}$.  To see that
$F_W^{\vec{y},\vec{u}}$ is also
satisfied by $\vec{v}$, simply repeat this argument above starting with
another
variable $W'$ in $\V - \vec{Y}$.  (Such a variable must exist since
$|\V| - |\vec{Y}|$ was assumed to be at least 2.)

It remains to show that $\vec{v}$ is the {\em unique\/}
solution to the equations in $T_{\vec{y}}(\vec{u})$.   Suppose there
were another
solution, say $\vec{v}'$, to the equations.  Suppose that for each
variable $W$ in
$\V - \vec{Y}$, the $W$ component of $\vec{v}'$ is
$w^{**}$. For some variable $Z$,
we must have $z^{**} \ne z^*$.  Since $Z_{\vec{y}}(\vec{u}) = z^*$, by
assumption, it follows from C1 that  $Z_{\vec{y}}(\vec{u})
\ne z^{**}$ is in $C$ (since $C$ is a maximal consistent set).
It is also easy to see that for each $W$ in $\V - \vec{Y}$, the vector
$\vec{v}'$
is also a solution to the equations in $T_{\vec{y}w^{**}}(\vec{u})$.
Let $W$ be a variable other than $Z$ in $\V - \vec{Y}$.  By the
induction hypothesis, it follows that $W_{\vec{y}z^{**}}(\vec{u}) =
w^{**}$ and
$Z_{\vec{y}w^{**}}(\vec{u}) = z^{**}$ are both in $C$.  By C5 (reversibility),
$Z_{\vec{y}}(\vec{u}) = z^{**}$ is in $C$.  But this contradicts the
consistency of $C$.

This completes the proof in the case of $\Tun(\S)$.  Essentially the
same proof
works for $\Trec$.  We just need to observe that C6 guarantees that the
theory we construct can be taken to be recursive.  To see this, given a
formula $\phi$ consistent with $\Trec$, consider a maximal set $C$ of
formulas consistent with $\Trec$ that contains $\phi$.  Let $T_C$ be the
causal model determined by $C$, as above.  The
set $C$ also determines a relation $\prec$ on the
exogenous variables: define $Y \prec Z$ if $Y \leadsto Z \in C$.
It easily follows from C6 that the transitive closure $\prec^*$ of
$\prec$ is a partial order: if $X \prec^* Y$ and $Y \prec^* X$, then $X
= Y$.  Any total order on the variables consistent $\prec^*$
gives an ordering for which $T_C$ is recursive.
\eprf

\medskip

\othm{thm2}
$\AXex$ (resp., $\AXunex$, $\AXrecex$) is a sound and complete
axiomatization for $\Lex(\S)$ with respect to $\T(\S)$ (resp.,
$\Tun(\S)$,
$\Trec(\S)$).
\eothm

\prf Soundness proceeds much as that of Theorem~\ref{thm1}; I leave
details
to the reader.  For completeness, we again proceed much as in the proof
of Theorem~\ref{thm1}.  Because the proofs are so similar in spirit, I
just sketch the proof for $\AXex$; the
modifications for $\AXunex$ and $\AXrecex$ are left to the reader.

Again, given a formula $\phi$ consist with $\AXex$, we consider a
maximal consistent set of formulas containing $\phi$ that is consistent
with $\AXex$, and use it to construct a causal model $T$.  Note that D9
suffices for this, because in defining $F_X(\vec{u},\vec{y})$, we
needed to know only the unique $x$ such that $[\vec{Y} \gets
\vec{y}](X(\vec{u}) = x)$ for $\vec{Y} = \V-X$, and D9 (together
with D1) assures
us that there is a unique such $x$.  Again, we want to show that all the
formulas in $C$ are satisfied by $T$.

To do this, it clearly suffices to show that for every formula $\psi$ of
the form
$\langle \vec{Y} \gets \vec{y} \phi$, we have $\psi$ in C iff $T \sat
\psi$.  We can reduce to considering even simpler formulas, namely, ones
where $\phi$ has the form $\vec{X}(\vec{u}) = \vec{x}$, by
applying some of the axioms.   To see this, first observe that standard
arguments of modal logic (using D0, D7, D8, and MP) show that $\langle
\vec{Y} \gets \vec{y}\>(\phi_1 \lor \phi_2)$ is provably equivalent to
$\<\vec{Y} \gets \vec{y}\>\phi_1 \lor
\<\vec{Y} \gets \vec{y}\>\phi_2$.  That means we can assume without
loss of generality that $\phi$ is a conjunction of formulas of the form
$X(\vec{u}) \gets x$ and their negations.   From D2 it follows that
$\<\vec{Y} \gets \vec{y}\>(\phi \land X(\vec{u}) \ne x)$ is equivalent
to $\<\vec{Y} \gets \vec{y}\>(\phi \land (\lor_{x' \in \R(X) -
\{x\}}X(\vec{u}) = x')$.  Thus, we can assume without loss of generality
that $\phi$ has no negations.  By applying D11, we can assume without
loss of generality that the same setting $\vec{u}$ of the exogenous
variables is used in all the conjuncts.  Thus, it suffices to show that
$\<\vec{Y} \gets \vec{y}\>(\vec{X}(\vec{u}) = \vec{x}) \in
C$ iff $T \sat \<\vec{Y} \gets \vec{y}\>(\vec{X}(\vec{u}) = \vec{x})$
for $\vec{X} = \V - \vec{Y}$.

To do this, we proceed by induction on
$|\V| -
|\vec{Y}|$ again.  The base case is dealt with using D4, as before.  So
assume that $k \ge 1$ and $|\V| - |\vec{Y}| = k+1$.  Suppose that
$\<\vec{Y} \gets \vec{y}\>(\vec{X}(\vec{u}) =\vec{x}) \in C$.
Let $X_1, X_2 \in \vec{X}$.  Suppose that $X_1 \gets x_1$ and $X_2 \gets
x_2$
are the assignments to $X_1$ and $X_2$ in $\vec{X} \gets \vec{x}$.   Let
$\vec{X}' \gets \vec{x}'$ and $\vec{X}'' \gets \vec{x}''$ be the
result of removing $X_1 \gets x_1$ and $X_2 \gets x_2$, respectively,
from $\vec{X} \gets \vec{x}$.  By D3, both
$\<\vec{Y} \gets \vec{y}; X_1 \gets x_1\>(\vec{X}''(\vec{u})
=\vec{x}'')$ and
$\<\vec{Y} \gets \vec{y}; X_2 \gets x_2\>(\vec{X}'(\vec{u}) =\vec{x}')$
are in $C$.  By the induction hypothesis, both of these formulas are
true in $T$.  By the soundness of D5, it follows that
$T \sat \<\vec{Y} \gets \vec{y}\>(\vec{X}(\vec{u}) =\vec{x}')$, as
desired.

Conversely, suppose that
$T \sat \<\vec{Y} \gets \vec{y}\>(\vec{X}(\vec{u}) =\vec{x}')$.  Then,
since D3 is sound, we have that
$T \sat
\<\vec{Y} \gets \vec{y}; X_1 \gets x_1\>(\vec{X}''(\vec{u})
=\vec{x}'')$ and
$T \sat \<\vec{Y} \gets \vec{y}; X_2 \gets x_2\>(\vec{X}'(\vec{u})
=\vec{x}')$.  By the induction hypothesis, we have that both
$\<\vec{Y} \gets \vec{y}; X_1 \gets x_1\>(\vec{X}''(\vec{u})
=\vec{x}'')$ and
$\<\vec{Y} \gets \vec{y}; X_2 \gets x_2\>(\vec{X}'(\vec{u}) =\vec{x}')$
are in $C$.  We now apply D5 to complete the proof. \eprf

\bigskip

\othm{finite1} A formula $\phi \in \Lex(\S)$ is satisfiable in
$\Trec(\S)$ (\respc $\Tun(\S)$) iff it is satisfiable in
$\Trec(\S_\phi)$ (\respc $\Tun(\S_\phi)$).  \eothm

\medskip

\prf Clearly, if a formula is satisfiable in $\Trec(\S_\phi)$ (\respc
$\Tun(\S_\phi)$), then it is satisfiable in $\Trec(\S)$ (\respc
$\Tun(\S)$). We can easily convert a causal model $T = (\S_\phi, \F) \in
\Trec(\S_\phi)$ satisfying $\phi$ to a causal model $T' = (\S,\F')
\in \Trec(\S)$ satisfying $\phi$ by simply defining $F_X'$ to be a
constant, independent of its arguments, for $X \in \V - \V_\phi$; if
$X \in \V_\phi$, define $F_X'(\vec{u},\vec{x},\vec{y}) =
F_X(\vec{u},\vec{x})$, where $\vec{u} \in \R(U^*)$,
$\vec{x} \in \times_{Y \in \V_\phi - \{X\}} \R(Y)$ and
$\vec{y} \in \times_{Y \in \V - \V_\phi}\R(Y)$; if $\vec{u} \notin
\R(U^*)$, define $F_X'(\vec{u},\vec{x},\vec{y})$ to be an arbitrary
constant.  An identical transformation works for $T \in \Tun(\S_\phi)$.

For the converse, suppose that $\phi$ is satisfiable
in a causal model $T = (\S,\F) \in \Trec(\S)$.
Thus, there is an ordering $\prec$ on the variables in $\V$ such that if
$X \prec Y$, then $F_X$ is independent of the value of $Y$.  This means
we
can view $F_X$ as a function of the exogenous variables in $\U$ and the
variables $Y \in \V$ such that $Y \prec X$.  Let $\Pre(X)
= \{Y \in \V: Y \prec X\}$.  For convenience, I allow
$F_X$ to take as arguments the values of
only the variables in $\U \union
\Pre(X)$, rather than requiring its arguments to include the values of
all the variables in $\U \union \V - \{X\}$.  Now define functions
$F_X': (\times_{U \in \U} \R(U))
\times (\times_{Y \in \V_\phi -\{X\}} \R(Y)) \rightarrow \R(X)$ for all $X
\in
\V$ by induction on $\prec$ (that is, start with the $\prec$-minimal
element, whose value is independent of that of all the other variables,
and work up the $\prec$ chains).  Suppose $X \in \V_\phi$ and $\vec{x}$
is a vector of values for the variables in $\V_\phi - \{X\}$.  If $X$ is
$\prec$-minimal, then define
$F'_{X}(\vec{u},\vec{x}) = F_{X}(\vec{u})$.  In general, define
$F'_{X}(\vec{u},\vec{x}) = F_X(\vec{u},\vec{z})$, where
$\vec{z}$ is a vector of values for the variables
in $\Pre(X)$ defined as follows.  If $Y \in \Pre(X) \inter \V_\phi$, then
the value of the $Y$ component
in $\vec{z}$ is the value of the $Y$ component in $\vec{y}$; if $Y
\in \Pre(X) - \V_\phi$, then the value of the $Y$ component in
$\vec{z}$ is $F'_Y(\vec{u},\vec{x})$.  (By the induction  hypothesis,
$F'_Y(\vec{u},\vec{x})$ has already been defined.)
Now define a causal model $T' = (\S_\phi, \F')$.  It
is easy to check that $T' \in \Trec(\S_\phi)$ (the ordering of the
variables is just $\prec$ restricted to $\V_\phi$).  Moreover, the
construction guarantees that if $\vec{X} \subseteq \V_\phi$, then
the solutions to the equations $T'_{\vec{X}
\gets \vec{x}}(\vec{u})$ and $T_{\vec{X} \gets \vec{x}}(\vec{u})$ are
the
same, when restricted to the variables in $\V_\phi$. It follows that $T'$
satisfies $\phi$.

The argument in the case that $T \in \Tun(\S)$ is similar in spirit.
For $X \in \V_\phi$, $\vec{u} \in (\times_{U \in \U} \R(U))$, and
$\vec{x} \in (\times_{Y \in \V_\phi - \{X\}} \R(Y))$, define
$F'_X(\vec{u},\vec{x})$ to be the value of $X$ in the unique
solution to the equations in $T_{\V_\phi - \{X\} \gets
\vec{x}}(\vec{u})$.%
\footnote{This definition is easily seen to agree with the earlier
definition of $F'_X$ if $T \in \Trec$.}
It is again straightforward to check that now $T' = (\S_\phi, \F')
\in \Tun(S_\phi)$ and satisfies $\phi$. \eprf

\bigskip

\othm{finite1a}
A formula $\phi \in \Lex(\S)$ is satisfiable in
$\T(\S)$ iff $\phi$ is satisfiable in $\T(\S_\phi^+)$.  \eothm

\medskip

\prf
If $||\S|| \le ||\S_\phi||^2 + ||\S_\phi||$ then the proof is immediate,
so suppose that $||\S|| > ||\S_\phi||^2 + ||\S_\phi||$
and $\phi$
is satisfied in a causal model $T = (\S,\F) \in \T(\S)$.
Before going on
with the proof, it is useful to define some notation.  Let $\V =
\{X_1, \ldots, X_m\}$, where $\V_\phi = \{X_1, \ldots, X_k\}$ and $\V -
\V_\phi = \{X_{k+1}, \ldots, X_m\}$.
Given a vector $\vec{x} \in \R(X^*) = \times_{X \in \V_\phi} \R(X)$
and $X_i \in \V_\phi$,
let $\vec{x}_{-i}$ denote the vector excluding the value for $X_i$.
For each $X_i \in \V_\phi$, choose two values $x_{i0}$ and $x_{i1}$ in
$\R(X_i)$.  Define $T' =
(\S_\phi, \F')$ by defining
$F_X'(\vec{u},\vec{x}_{-i},\vec{y}_{-i},y_i) =
x$, where
\begin{itemize}
\item $x = y_i$ if $\vec{x}_{-i} = \vec{y}_{-i}$ and $X=y_i$ in some
solution
to the equations in $T_{\V_\phi - \{X_i\} \gets \vec{x}_{-i}}(\vec{u})$;
\item $x = x_{i0}$ if $y_i \ne x_{i0}$ and either
$\vec{x}_{-i} \ne  \vec{y}_{-i}$ or there is no solution to
the equations in $T_{\V_\phi - \{X\} \gets \vec{x}_{-i}}(\vec{u})$ in
which
$X = y_i$;
\item $x = x_{i1}$ otherwise.
\end{itemize}
Finally, define $F_{X^*}(\vec{u},\vec{x}) = \vec{x}$.

I now show that the construction again
guarantees that if $\vec{X} \subseteq
\V_\phi$, then the solutions to the equations $T'_{\vec{X}
\gets \vec{x}}(\vec{u})$ and $T_{\vec{X} \gets \vec{x}}(\vec{u})$ are the
same, when restricted to the variables in $\V_\phi$.  First suppose that
$(\vec{y},\vec{z})$ is a solution to the equations in
$T_{\vec{X} \gets \vec{x}}(\vec{u})$, where $\vec{y} \in \R(X^*)$
and $\vec{z} \in \times_{Y \in \V - \V_\phi} \R(Y)$.
It must be the case that
$\vec{x}$ and $\vec{y}$ agree on the variables in $\vec{X}$, so
$(\vec{y},\vec{z})$ is
also a solution of the equations in $T_{\V_\phi - \{X_i\} \gets
\vec{y}_{-i}}(\vec{u})$ if $X_i \in \V_\phi - \vec{X}$.  Thus,
$F'_{X_i}(\vec{u},\vec{y}_{-i},\vec{y})
= y_i$.  It follows that $(\vec{y},\vec{y})$ is a solution to the
equations in $T'_{\vec{X} \gets \vec{x}}(\vec{u})$.

Conversely, suppose that $(\vec{y},\vec{y}')$
is a solution to the
equations in $T'_{\vec{X} \gets \vec{x}}(\vec{u})$.  Then the
definition of $F_{X^*}$ guarantees that
$\vec{y} = \vec{y}'$.  Moreover, since $\vec{x}$ and
$\vec{y}$ agree on the variables in $\vec{X}$, $(\vec{y},\vec{y})$
must also be a solution to the equations in $T'_{\V_\phi - \{X_1\}
\gets \vec{y}_{-1}}(\vec{u})$.  Thus,
$F'_{X_1}(\vec{u},\vec{y}_{-1},\vec{y}) = y_1$, which means that there
must be some vector $\vec{z}$ of values for the variables in $\V
- \V_\phi$ such that $(\vec{y},\vec{z})$ is a solution to the equations
in $T_{\V_\phi - \{X_1\} \gets \vec{y}_{-1}}(\vec{u})$.  But then it is
easy to check that $(\vec{y},\vec{z})$ must in fact be a solution
to the equations
in $T_{\V_\phi - \{X_i\} \gets \vec{y}_{-i}}(\vec{u})$ for all $i = 1,
\dots, k$.  It follows that $(\vec{y},\vec{z})$ is a solution to the
equations in $T_{\vec{X} \gets \vec{x}}(\vec{u})$, as desired.  This
suffices to prove this direction of the theorem.

Now suppose that $\phi$ is satisfied in
a causal model $T = (\S_\phi^+,\F) \in
\T(\S_\phi^+)$.  Since $||\S|| > ||\S_\phi||^2 +||\S_\phi||$,
there must be an injective function $f: \R(X^*)
\rightarrow \times_{Y \in \V - \V_\phi} \R(Y)$ and
two distinct vectors
$\vec{y}_0 = (y_{01}, \ldots, y_{0k}), \,  \vec{y}_1 = (y_{11},
\ldots, y_{1k})$ that are not in the range of $f$.
Choose two distinct vectors
$\vec{x}_0 = (x_{10}, \ldots, x_{k0}), \,  \vec{x}_1 = (x_{11},
\ldots, x_{k1}) \in \R(X^*)$.
Define $T' = (\S,\F') \in \T(\S)$
as follows. If $X_i \in \V_\phi$, $\vec{x}_{-i} \in \times_{Y \in
\V_\phi - \{X_i\}} \R(Y)$, $\vec{z} \in \R(X^*)$, and
$\vec{y} \times_{Y \in \V - \V_\phi} \R(Y)$,
let
$$F'_{X_i}(\vec{x}_{-i},\vec{y}) =
\left\{
\begin{array}{ll}
F_{X_i}(\vec{x}_{-i},\vec{z}) &\mbox{if $f(\vec{z}) = \vec{y}$,}\\
x_{0i} &\mbox{if $\vec{y}$ is not in the range of $f$, $\vec{y} \ne
\vec{y}_1$,}\\
x_{1i} &\mbox{otherwise.}
\end{array}\right.
$$
If $X_j \in \V - \V_\phi$, $\vec{x} \in \R(X^*)$
and $\vec{y}_{-j}  \in \times_{Y \in \V - \V_\phi - \{X_j\}} \R(Y)$, then
let
$$F_{X_j}'(\vec{x},\vec{y}_{-j}) =
\left\{
\begin{array}{ll}
y \ \ \ &\mbox{if $f(F_{X^*}(\vec{x})) = (\vec{y}_{-j},y)$,}\\
y_{0j} &\mbox{if $f(F_{X^*}(\vec{x})) \ne (\vec{y}_{-j},y')$ for all $y'
\in \R(X_j)$, $\vec{x} \ne \vec{x_0}$,}\\
y_{1j} &\mbox{otherwise.}
\end{array}\right.
$$

Again, I show that the construction
guarantees that if $\vec{X} \subseteq
\V_\phi$, then the solutions to the equations $T'_{\vec{X}
\gets \vec{x}}(\vec{u})$ and $T_{\vec{X} \gets \vec{x}}(\vec{u})$ are the
same, when restricted to the variables in $\V_\phi$.  First suppose that
$(\vec{y},\vec{z})$ is a solution to the equations in
$T_{\vec{X} \gets \vec{x}}(\vec{u})$, where $\vec{y}, \vec{z} \in
\R(X^*)$.  It is easy to check that $(\vec{y},f(\vec{z}))$ is a solution
to the equations in $T'_{\vec{X} \gets \vec{x}}(\vec{u})$.
Conversely, suppose that
$(\vec{y},\vec{z})$ is a solution to the equations in
$T'_{\vec{X} \gets \vec{x}}(\vec{u})$, where $\vec{y} \in
\R(X^*)$ and $\vec{z} \in \times_{Y \in \V - \V_\phi} \R(Y)$.
I claim that we must have $\vec{z} = f(F_{X^*}(\vec{y}))$.
If, in fact, this is the case, then it is easy to check that
$(\vec{y},F_{X^*}(\vec{y})$ is a solution to the equations in
$T_{\vec{X} \gets \vec{x}}(\vec{u})$.  On the other hand, if
$\vec{z} \ne f(F_{X^*}(\vec{y}))$,
then the definition of $F_{X_j}'$ for $X_j \in \V - \V_\phi$
guarantees that $\vec{z} = \vec{y}_0$ unless $\vec{y} = \vec{x}_0$; if
$\vec{y} = \vec{x}_0$, then $\vec{z} = \vec{y}_1$.  But the definition
of $F_{X_i}$ for $X_i \in \V_\phi$ guarantees that if
$\vec{z} = \vec{y}_0$, then $\vec{y} = \vec{x}_0$: otherwise, $\vec{y}
= \vec{x}_1$.  Thus, $(\vec{y},\vec{z})$ is a solution iff $\vec{z} =
f(F_{X^*}(\vec{y}))$.  This suffices to prove the result.  \eprf

\bigskip

\othm{decprob2}  Given as input a pair $(\phi,\S)$, where $\phi
\in \Lex(\S)$ (\respc $\Lprop(\S)$) and $\S$ is a finite
signature, the problem of deciding
if $\phi$ is satisfiable with respect to $\Trec(\S)$ (\respc $\Tun(\S)$,
$\T(\S)$) is NP-complete (\respc NP-hard) in $|\phi|$;
if $\phi \in \LGP(\S)$, then the problem of deciding if $\phi$ is
satisfiable in $\Trec(\S)$ (\respc $\Tun(\S)$) is NP-complete (\respc
NP-hard).
\eothm

\medskip

\prf The NP-lower bound is easy for $\Lex(\S)$ and $\Lprop(\S)$, since
there is an obvious way to encode the satisfiability problem
for propositional logic into the satisfiability problem for $\Lex$ and
$\Lprop$.  Given a propositional formula $\phi$
with primitive propositions $p_1,
\ldots, p_k$, let $\S = (\emptyset, \{X_1, \ldots, X_k\}, \R)$,
where $\R(X_i) = \{0,1\}$ for $i = 1, \ldots, k$.
Replace each occurrence of the primitive proposition $p_i$ in $\phi$
with the formula $X_i = 1$.  This gives us a formula $\phi'$
in $\Lprop(\S)$.
It is easy to see that if $\phi'$ is satisfiable in a causal model $T
\in \T(\S)$
(and, {\em a fortiori\/} if $\phi'$ is satisfiable in a causal model $T$
in either $\Trec(\S)$ or $\Tun(\S)$) then the solution to the equations
in $T$ defines a satisfying assignment for $\phi$.  Conversely, if
$\phi$ is
satisfiable, say by some truth assignment $v$, then we can trivially
construct a causal model $T \in
\Trec(\S)$ such that $F_{X_i} = v(p_i)$.  (For simplicity, I assume that
valuations assign values 0 and 1 rather than {\bf false} and {\bf
true}.)

This trivial construction of $\phi'$
will not work for $\LGP(\S)$, since we do not have disjunctions or
negations
available.  The lack of negations does not cause a problem.  We can
assume without loss of generality that the negations occur only in front
of primitive propositions, and we can capture $\neg p_i$ by the formula
$X_i = 0$.  The idea for dealing with disjunctions is that a formula
such as $p_1 \lor \neg p_2 \lor p_3$ is translated to $[X_1 \gets 0; X_2
\gets 1; X_3 \gets 1](Y=0)$, where $Y$ is a fresh variable.
Essentially, we are viewing $p_1 \lor \neg p_2 \lor p_3$
as $(\neg p_1 \land p_2 \land \neg p_3) \rimp \false$, which is why we
write, for example,
$X_1 \gets 0$ even though $p_1$ appears positively in the disjunction.

To make matters simpler,
assume that $\phi$ is a formula in 3-CNF.  This suffices for
NP-hardness, since the satisfiability problem for 3-CNF formulas is also
NP-hard \cite{GarJoh}.  Suppose $\phi$ is of the form $c_1 \land \ldots
\land c_m$, where each $c_l$ is a {\em clause\/} consisting of a
disjunction of three primitive propositions and their negations.
Suppose that the
primitive propositions that appear in $\phi$ are $p_1, \ldots, p_k$.
Let $\S = (\emptyset,\{X_1, \ldots, X_k,
Y_1, \ldots, Y_m\}, \R)$, where
$\R(X_{i}) = \R(Y_j) = \{0,1\}$ for all $i, j$.  Suppose
that $c_j$, the $j$th clause of $\phi$, is of the form $q_{j1} \lor
q_{j2} \lor q_{j3}$,
where $q_{ji}$ is either
$p_{j_i}$ or $\neg p_{j_i}$ for some $j_i$. Let
$c_j^t$ be the $\LGP$ formula
$$[X_{j_1} \gets x_{j1}; X_{j_2}
\gets x_{j2}; X_{j_3} \gets x_{j3}](Y_j = 0),$$
where $x_{jh}$ is 0 if $q_{jh}$ is $p_{j_h}$
and $x_{jh}$ is 1 if $q_{jh}$ is $\neg p_{j_h}$ for $h = 1,2,3$.
Let $\phi'$ be $$[ \true] (Y_1 = 1 \land \ldots \land Y_m = 1)
\land c_1^t \land \ldots \land c_m^t.$$

I claim that $\phi$ is a satisfiable propositional formula iff the
$\LGP$ formula
$\phi'$ is satisfiable in $\Trec(\S)$ (\resp $\Tun(\S)$).
First suppose that $\phi'$ is
satisfiable, say in some model $T \in \Tun(\S)$.  (If this
direction holds for
$T \in \Tun(\S)$, it clearly holds {\em a fortiori\/} for $T \in
\Trec(\S)$.)  Let
$\vec{z}$ be the unique solution to the equations in $T$.  By
construction, the
$Y_j$ component of $\vec{z}$ is 1 for $j = 1,
\ldots, m$.  Let $x^*_{i}$ be the value of the $X_{i}$ component in
$\vec{z}$.
Consider the valuation $v$ such that $v(p_i) = x^*_{i}$.
I claim that $v(\phi) = 1$.  To see this, suppose that clause
$c_j$ is $q_{j1} \lor q_{j2} \lor q_{j3}$.  If $v$ makes $q_{j1}$,
$q_{j2}$, and $q_{j3}$ false, then we must have $x_{jh} =
x^*_{j_h}$ for $h=1,2,3$.
Since $T \sat [X_{j_1} \gets x_{j1}; X_{j_2}
\gets x_{j2}; X_{j_3} = x_{j3})](Y_j = 0)$ and the value of the
$X_{j_h}$ component of $\vec{z}$ is $x_{jh}$ for $h = 1, 2, 3$, it
follows that $\vec{z}$ is a solution to the equations in
$T_{X_{j_1} \gets x_{j1}; X_{j_2}
\gets x_{j2}; X_{j_3} \gets x_{j3}}$.  But this contradicts the fact
that $T \sat [X_{j_1} \gets x_{j1}; X_{j_2}
\gets x_{j2}; X_{j_3} \gets x_{j3}](Y_j = 0)$ (since the $Y_j$ component
of
$\vec{z}$ is 1).  It
follows that $v(c_j) = v(q_{j1} \lor q_{j2} \lor q_{j3}) = 1$.
Since this is true
for all clauses $c_j$, we must have that $v(\phi) = 1$.

For the converse, suppose that $\phi$ is satisfiable, say by valuation
$v$.  I show that
$\phi'$ is satisfiable in $T \in \Trec(\S)$.  Order the variables so
that $X_{j_1}, X_{j_2}, X_{j_3} \prec Y_j$.
(There are many orderings of the variables that satisfy these
constraints; any one will do.)  Define $F_{X_i} = v(p_i)$ (so that
$F_{X_i}$ is a constant, independent of its arguments);  define
$F_{Y_j}(x_{j_1}, x_{j_2}, x_{j_3}) = 1$ if $(x_{j_1},x_{j_2},x_{j_3})
= (v(p_{j_1}),v(p_{j_2}),v(p_{j_3}))$ and 0 otherwise.  It is easy to
check that $T \sat \phi'$, as desired.

For the NP upper bound in the case of $\Trec(\S)$, it clearly suffices
to deal with $\phi \in
\Lex$. Suppose we are given $(\phi,\S)$ with $\phi \in \Lex$.
We want to check if $\phi$ is
satisfiable in $\Trec(\S)$.  The basic
idea in to guess a causal model $T$
and verify that
it indeed satisfies $\phi$.  There is a problem with this though.  To
completely describe a model $T$, we need to describe the functions
$F_X$.  However, there may be many variables $X$ in $\S$ and they can
have many possible inputs.  Just describing these functions may take
time much longer than polynomial in $\phi$.  Part of the solution to
this problem is provided by Theorem~\ref{finite1},
which tells us that
it suffices to check whether $\phi$ is satisfiable in $\Trec(\S_\phi)$.
In light of this, for the remainder of this part of the proof, I assume
without loss of generality
that $\S= \S_\phi$.
This limits the number of
variables that we must consider to $O(|\phi|)$.
But
even this does not solve our problem completely.  Since we are not given
any bounds on $|\R(Y)|$ for variables $Y$ in $\S_\phi$, even
describing the functions $F_Y$ for the variables $Y$ that appear in
$\phi$ on all their possible input vectors could take time
much more than polynomial in $\phi$.  The solution is to give only a
short partial description of a model $T$ and show that this suffices.
Consider all pairs $(\vec{Y} \gets \vec{y}, \vec{u})$ such
that there is a subformula of $\phi$ of the
form $[\vec{Y} \gets \vec{y}] \psi$ and $\vec{u}$ appears in $\psi$.
Let $R$ be the set of all such pairs.  Note that $|R| < |\phi|^2$.
We say that two causal models
$T$ and $T'$ in $\Trec(\S)$ {\em agree on $R$\/} if, for all pairs
$(\vec{Y} \gets \vec{y}, \vec{u}) \in R$, the (unique) solutions to
the equations in $T_{\vec{Y} \gets \vec{y}}(\vec{u})$ and
$T'_{\vec{Y} \gets \vec{y}}(\vec{u})$ are the same.  It is easy to see
that if $T$ and $T'$ agree on $R$, then either both $T$ and $T'$
satisfy
$\phi$ or neither do.  That is, all we need to know about a causal
model is how it deals with the {\em relevant equations}---those
corresponding to pairs in $R$.

For each
pair $(\vec{Y} \gets \vec{y}, \vec{u})
\in R$, guess a vector $\vec{v}(\vec{Y} \gets \vec{y},\vec{u})$ of
values for the endogenous variables; intuitively, these are the unique
solutions to the relevant equations in a model satisfying $T$.
Given these guesses, it is easy
to check if $\phi$ is satisfied in a model where these guesses do indeed
represent the solutions to the relevant equations.  It remains to show
that there exists a causal model in $\Trec(\S)$
where the relevant equations have these solutions.
To do this,
first guess an ordering $\prec$ on the variables.  We can then
verify, for each fixed $\vec{u}$ that appears in $\phi$, whether the
solution vectors $\vec{v}(\vec{Y} \gets \vec{y},\vec{u})$
guessed for the relevant equations are
{\em compatible with $\prec$}, in the sense
that it is not the case that there are two solutions $(\vec{u},\vec{x})$
and $(\vec{u},\vec{x}')$ such that some variable $X$ takes on different
values in $\vec{x}$ and $\vec{x}'$, but all variables $Y$ such that $Y
\prec X$ take on the same values in $\vec{x}$ and $\vec{x}'$.  It is
easy to see that if the solutions are compatible with $\prec$, we can
define the functions $F_X$ for $X \in \V$ such that all the equations
hold and
$F_X$ is independent of the values of $Y$ if $X \prec Y$ for all $X, Y
\in \V$.  (Note we never actually have to write out the functions $F_X$,
which may take too long; we just have to know they exist.)  To
summarize, as long as we can guess some solutions to the
relevant equations such that a causal model that has these
solutions satisfies
$\phi$, and an ordering $\prec$ such that these solutions are compatible
with $\prec$, then $\phi$ is satisfiable in $\Trec(\S)$.  Conversely, if
$\phi$ is satisfiable in $T \in \Trec(\S)$, then there clearly are
solutions to the relevant equations that satisfy $\phi$ and an ordering
$\prec$ such that
these solutions are compatible with $\prec$.
(We just take the solutions and the ordering $\prec$ from $T$.)
This shows that the satisfiability problem for $\Trec$ is in NP, as
desired.
\eprf

\bibliographystyle{theapa}
\bibliography{z,joe,expl}

\begin{thebibliography}{}

\bibitem[\protect\BCAY{Chajewska \BBA\ Halpern}{Chajewska \BBA\
  Halpern}{1997}]{CH97}
Chajewska, U.\BBACOMMA\  \BBA\ Halpern, J.~Y. \BBOP1997\BBCP.
\newblock \BBOQ Defining explanation in probabilistic systems\BBCQ\
\newblock In {\Bem Proc.~Thirteenth Conference on Uncertainty in Artificial
  Intelligence (UAI '97)}, \BPGS\ 62--71.

\bibitem[\protect\BCAY{Druzdzel \BBA\ Simon}{Druzdzel \BBA\
  Simon}{1993}]{DruzdzelSimon93}
Druzdzel, M.~J.\BBACOMMA\  \BBA\ Simon, H.~A. \BBOP1993\BBCP.
\newblock \BBOQ Causality in bayesian belief networks\BBCQ\
\newblock In {\Bem Uncertainty in Artificial Intelligence~9}, \BPGS\ 3--11.

\bibitem[\protect\BCAY{Galles \BBA\ Pearl}{Galles \BBA\
  Pearl}{1997}]{GallesPearl97}
Galles, D.\BBACOMMA\  \BBA\ Pearl, J. \BBOP1997\BBCP.
\newblock \BBOQ Axioms of causal relevance\BBCQ\
\newblock {\Bem Artificial Intelligence}, {\Bem 97\/}(1--2), 9--43.

\bibitem[\protect\BCAY{Galles \BBA\ Pearl}{Galles \BBA\
  Pearl}{1998}]{GallesPearl98}
Galles, D.\BBACOMMA\  \BBA\ Pearl, J. \BBOP1998\BBCP.
\newblock \BBOQ An axiomatic characterization of causal counterfactuals\BBCQ\
\newblock {\Bem Foundation of Science}, {\Bem 3\/}(1), 151--182.

\bibitem[\protect\BCAY{Garey \BBA\ Johnson}{Garey \BBA\ Johnson}{1979}]{GarJoh}
Garey, M.\BBACOMMA\  \BBA\ Johnson, D.~S. \BBOP1979\BBCP.
\newblock {\Bem Computers and Intractability: A Guide to the Theory of
  {NP}-completeness}.
\newblock W. Freeman and Co., San Francisco, Calif.

\bibitem[\protect\BCAY{Goldberger}{Goldberger}{1972}]{Goldberger72}
Goldberger, A.~S. \BBOP1972\BBCP.
\newblock \BBOQ Structural equation methods in the social sciences\BBCQ\
\newblock {\Bem Econometrica}, {\Bem 40\/}(6), 979--1001.

\bibitem[\protect\BCAY{Harel}{Harel}{1979}]{Har}
Harel, D. \BBOP1979\BBCP.
\newblock {\Bem First-Order Dynamic Logic}.
\newblock Lecture Notes in Computer Science, Vol. 68. Springer-Verlag,
  Berlin/New York.

\bibitem[\protect\BCAY{Heckerman \BBA\ Shachter}{Heckerman \BBA\
  Shachter}{1995}]{HeckShac}
Heckerman, D.\BBACOMMA\  \BBA\ Shachter, R. \BBOP1995\BBCP.
\newblock \BBOQ Decision-theoretic foundations for causal reasoning\BBCQ\
\newblock {\Bem Journal of Artificial Intelligence Research}, {\Bem 3},
  405--430.

\bibitem[\protect\BCAY{Henrion \BBA\ Druzdzel}{Henrion \BBA\
  Druzdzel}{1990}]{DH90}
Henrion, M.\BBACOMMA\  \BBA\ Druzdzel, M.~J. \BBOP1990\BBCP.
\newblock \BBOQ Qualitative propagation and scenario-based approaches to
  explanation of probabilistic reasoning\BBCQ\
\newblock In {\Bem Uncertainty in Artificial Intelligence~6}, \BPGS\ 17--32.

\bibitem[\protect\BCAY{Pearl}{Pearl}{1995}]{Pearl.Biometrika}
Pearl, J. \BBOP1995\BBCP.
\newblock \BBOQ Causal diagrams for empirical research\BBCQ\
\newblock {\Bem Biometrika}, {\Bem 82\/}(4), 669--710.

\bibitem[\protect\BCAY{Pearl}{Pearl}{1999}]{pearl:99}
Pearl, J. \BBOP1999\BBCP.
\newblock {\Bem Causality}.
\newblock Cambridge University Press, New York.
\newblock Forthcoming.

\bibitem[\protect\BCAY{Pearl \BBA\ Verma}{Pearl \BBA\
  Verma}{1991}]{PearlVerma91}
Pearl, J.\BBACOMMA\  \BBA\ Verma, T. \BBOP1991\BBCP.
\newblock \BBOQ A theory of inferred causation\BBCQ\
\newblock In {\Bem Principles of Knowledge Representation and Reasoning:
  Proc.~Second International Conference (KR '91)}, \BPGS\ 441--452.

\bibitem[\protect\BCAY{Spirtes, Glymour, \BBA\ Scheines}{Spirtes
  et~al.}{1993}]{SpirtesSG}
Spirtes, P., Glymour, C., \BBA\ Scheines, R. \BBOP1993\BBCP.
\newblock {\Bem Causation, Prediction, and Search}.
\newblock Springer-Verlag, New York.

\bibitem[\protect\BCAY{Strotz \BBA\ Wold}{Strotz \BBA\ Wold}{1960}]{SW60}
Strotz, R.~H.\BBACOMMA\  \BBA\ Wold, H. O.~A. \BBOP1960\BBCP.
\newblock \BBOQ Recursive vs.~nonrecursive systems: an attempt at
  synthesis\BBCQ\
\newblock {\Bem Econometrica}, {\Bem 28\/}(2), 417--427.

\end{thebibliography}
\end{document}